\documentclass[runningheads]{llncs}
\usepackage{graphicx}
\usepackage{comment}
\usepackage{amsmath,amssymb} 
\usepackage{color}
\usepackage{epsfig}
\usepackage{amsmath}
\usepackage{amssymb}
\usepackage{adjustbox} 
\usepackage[]{algorithm2e}
\usepackage{etoolbox}
\usepackage{multirow}
\usepackage{epigraph}
\usepackage{gensymb}
\usepackage{gensymb}
\usepackage{arydshln}
\usepackage{pifont}
\newcommand{\cmark}{\ding{51}}%
\usepackage[capbesideposition={inside,right},capbesidesep=quad]{floatrow}

\usepackage[utf8x]{inputenc}
\usepackage{pgfplots}
\usepackage{pgfplotstable}
\pgfplotsset{compat=newest}
\usepackage{tikz}
\usetikzlibrary{backgrounds}
\usepackage{color}
\usepackage{booktabs,siunitx} 
\usepackage{makecell}
\usepackage{enumitem}
\usepackage[caption=false]{subfig}
\usepackage[noadjust]{cite}
\usepackage{array,multirow,graphicx}
\usepackage{rotating}
 
 \usepackage{hyperref}
\hypersetup{
    colorlinks=true,
    linkcolor=blue,
    filecolor=magenta,      
    urlcolor=cyan,
}
 
\newfloatcommand{capbtabbox}{table}[][\FBwidth]
\usepackage[width=122mm,left=12mm,paperwidth=146mm,height=193mm,top=12mm,paperheight=217mm]{geometry}

\makeatletter
\@namedef{ver@everyshi.sty}{}
\makeatother

\begin{document}
\pagestyle{headings}
\mainmatter
\def\ECCVSubNumber{2632}  

\title{Semantic Object Prediction and Spatial Sound Super-Resolution with Binaural Sounds}


\titlerunning{Semantic Object Prediction with Binaural Sounds}
%
\author{Arun Balajee Vasudevan\inst{1} \and
Dengxin Dai\inst{1} \and
Luc Van Gool\inst{1,2}}
\authorrunning{Arun et al.}
%
\institute{ Computer Vision Lab, ETH Zurich \qquad \and KU Leuven \\
\email{\{arunv,dai,vangool\}@vision.ee.ethz.ch}}
\maketitle

\begin{abstract}
Humans can robustly recognize and localize objects by integrating visual and auditory cues. While machines are able to do the same now with images, less work has been done with sounds. This work develops an approach for dense semantic labelling of sound-making objects, purely based on binaural sounds. We propose a novel sensor setup and record a new audio-visual dataset of street scenes with eight professional binaural microphones and a 360\degree camera. 
The co-existence of visual and audio cues is leveraged for supervision transfer. In particular, we employ a cross-modal distillation framework that consists of a vision `teacher' method and a sound `student' method -- the student method is trained to generate the same results as the teacher method. This way, the auditory system can be trained without using human annotations. We also propose two auxiliary tasks namely, a) a novel task on Spatial Sound Super-resolution to increase the spatial resolution of sounds, and b) dense depth prediction of the scene. We then formulate the three tasks into one end-to-end trainable multi-tasking network aiming to boost the overall performance.
Experimental results on the dataset show that 1) our method achieves promising results for semantic prediction and the two auxiliary tasks; and 2) the three tasks are mutually beneficial -- training them together achieves the best performance and 3) the number and orientations of microphones are both important. 
The data and code will be released to facilitate the research in this new direction. Please refer to \href{https://www.trace.ethz.ch/publications/2020/sound_perception/index.html}{our project page}

\end{abstract}

\section{Introduction}


Autonomous vehicles and other intelligent robots will have a substantial impact on people’s daily life, both personally and professionally. While great progress has been made in the past years with visual perception systems~\cite{AD:Boss:08,Hecker_2018_ECCV,state:future:rescue:robotics:19}, we argue that auditory perception and sound processing also play a crucial role in this context~\cite{soundscape:icra17}. As known, animals such as bats, dolphins, and some birds have specialized on ``hearing" their environment. To some extent, humans are able to do the same -- to ``hear" the shape, distance, and density of objects around us~\cite{echolocating:listeners:00}. 
In fact, humans surely need to use this capability for many daily activities such as for driving -- certain alerting stimuli, such as horns of cars and sirens of ambulances, police cars and fire trucks, are meant to be heard, i.e. are primarily acoustic~\cite{soundscape:icra17}. 
Using multi-sensory information is a fundamental capability that allows humans to interact with the physical environment efficiently and robustly~\cite{Ernst2004MergingTS,merging:of:sense,seeing:hearing:actions:19}. Future intelligent robots are expected to have the same perception capability to be robust and to be able to interact with humans naturally. Furthermore, auditory perception can be used to localize  common objects like a running car, which is especially useful when visual perception fails due to adverse visual conditions or occlusions. 



Numerous interesting tasks have been defined at the intersection of visual and auditory sensing like sound source localization in images~\cite{sound:pixels:eccv18}, scene-aware audio generation for VR~\cite{scene-aware-audio}, geometry estimation for rooms using sound echos~\cite{room:geometry:acoustic:response:12}, sound source separation using videos~\cite{separate:sound:watching:video:18}, and scene~\cite{aytar2016soundnet} and object~\cite{vehicle:tracking:sound:iccv19} recognition using audio cues. There are also works to learn the correlation of visual objects and sounds~\cite{sound:pixels:eccv18,sounds:motion:iccv19}. While great achievements have been made, previous methods mostly focus on specific objects, e.g. musical instruments or noise-free rooms, or on individual tasks only, e.g. sound localization or geometry estimation. This work aims to learn auditory semantic prediction, for general, sound-making objects like cars, trains, and motorcycles in unconstrained, noisy environments.  

This work primarily focuses on semantic object prediction based on binaural sounds aiming to replicate human auditory capabilities. To enhance the semantic prediction, this work proposes two auxiliary tasks: depth prediction from binaural sounds and spatial sound super-resolution (S$^{3}$R). Studies~\cite{wang2015towards,mousavian2016joint} have shown that depth estimation and semantic prediction are correlated and are mutually beneficial. S$^{3}$R is a novel task aiming to increase the directional resolution of audio signals, e.g. from Stereo Audio to Surround Audio. S$^3$R, as an auxiliary task, is motivated from the studies~\cite{sound:localization:head:movement:14,sound:localization:head:movement:67} showing that humans are able to better localize the sounding sources by changing their head orientations. S$^{3}$R is also a standalone contribution and has its own applications. For instance, spatially resolved sounds improve the spatial hearing effects in AR/VR applications.
It offers better environmental perception for users and reduces the ambiguities of sound source localization~\cite{huang2012human}.

\begin{figure}[t]
\begin{tabular}{cccc}
  \centering \hspace{-4mm}
  \includegraphics[trim=150 350 150 300,clip,width=0.24\textwidth]{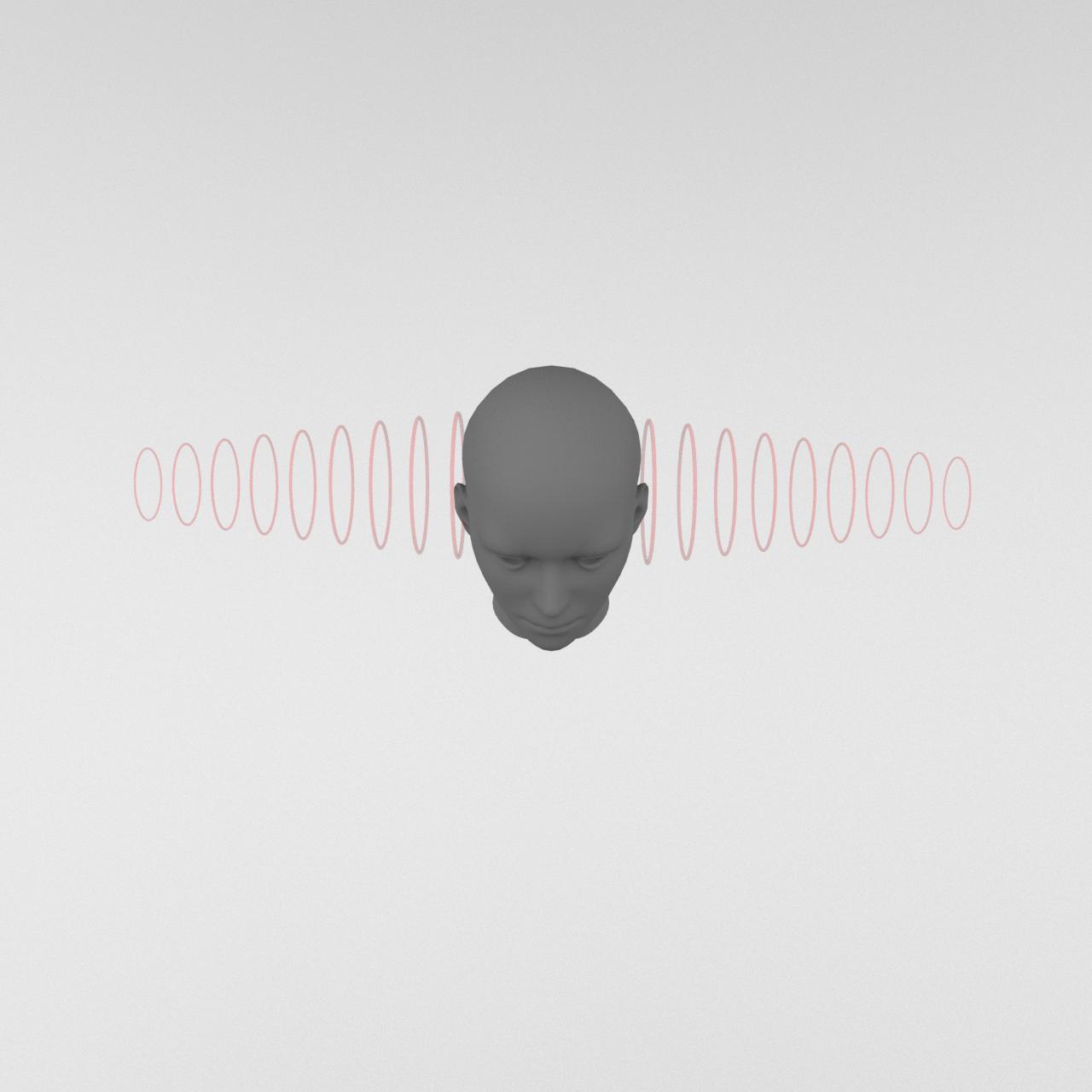} & \hspace{-3mm}
  \includegraphics[trim=150 350 150 300,clip,width=0.24\textwidth]{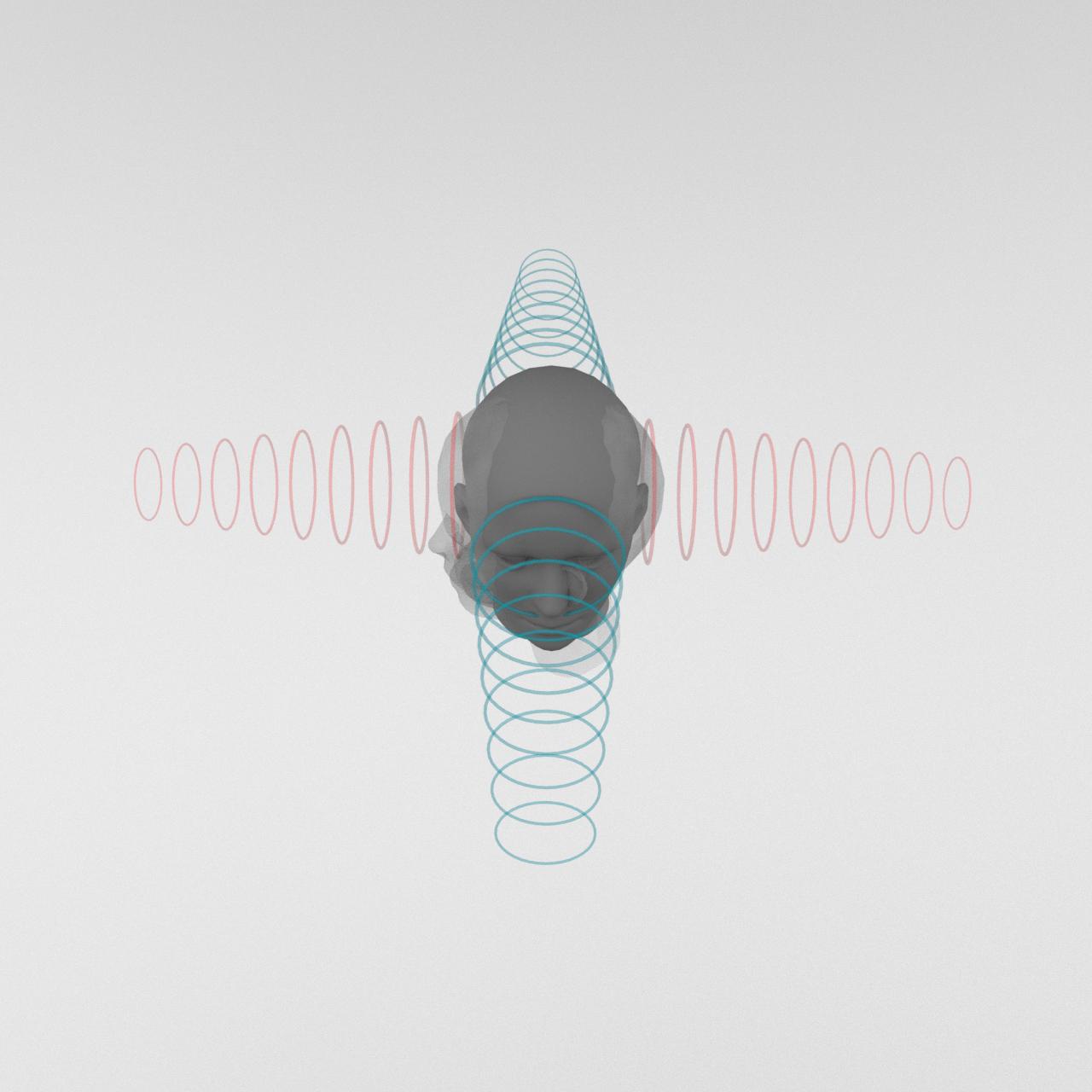} & \hspace{-3mm}
  \includegraphics[width=0.24\textwidth,height=0.155\textwidth]{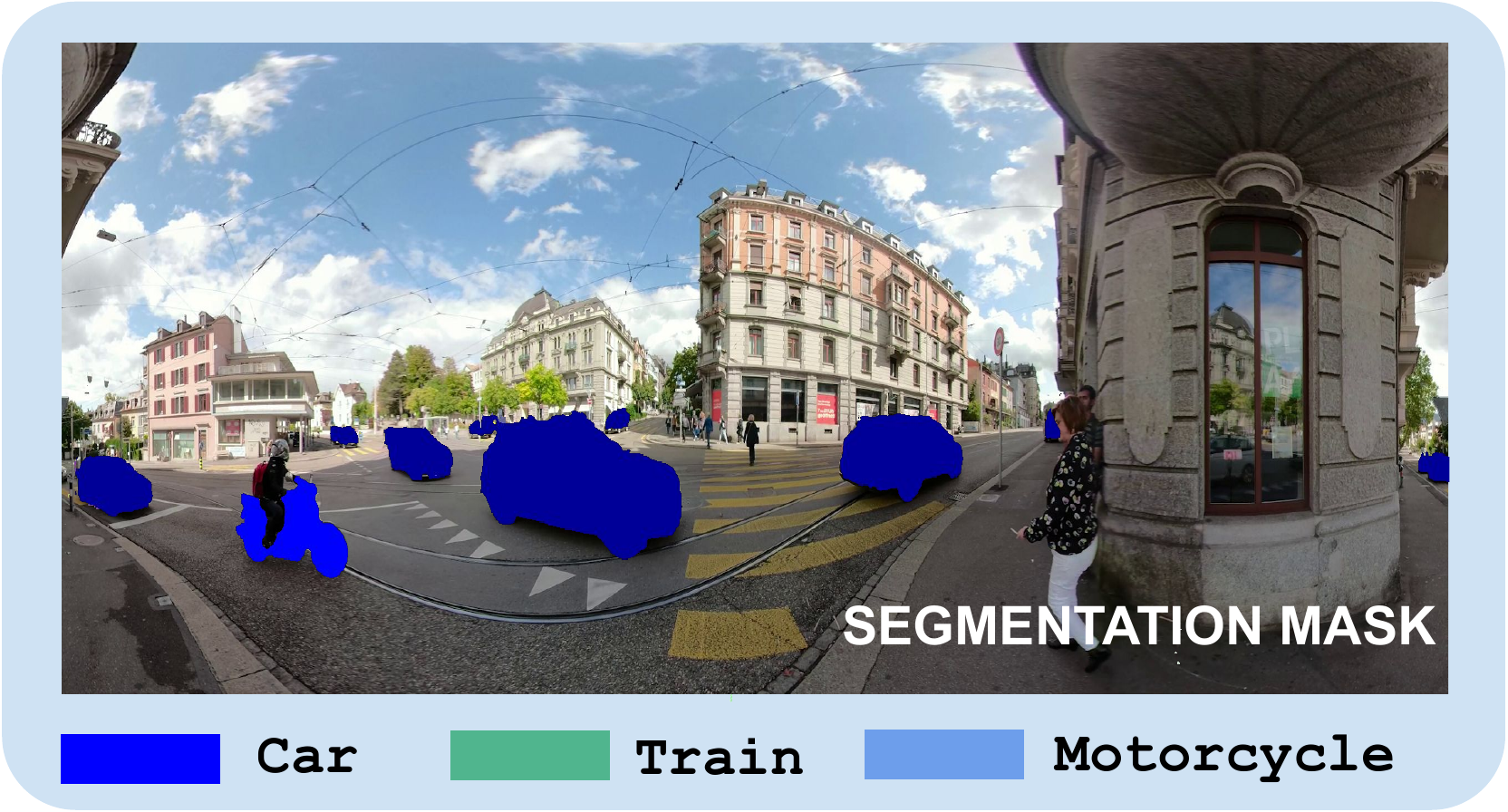} & 
  \includegraphics[width=0.24\textwidth,height=0.155\textwidth]{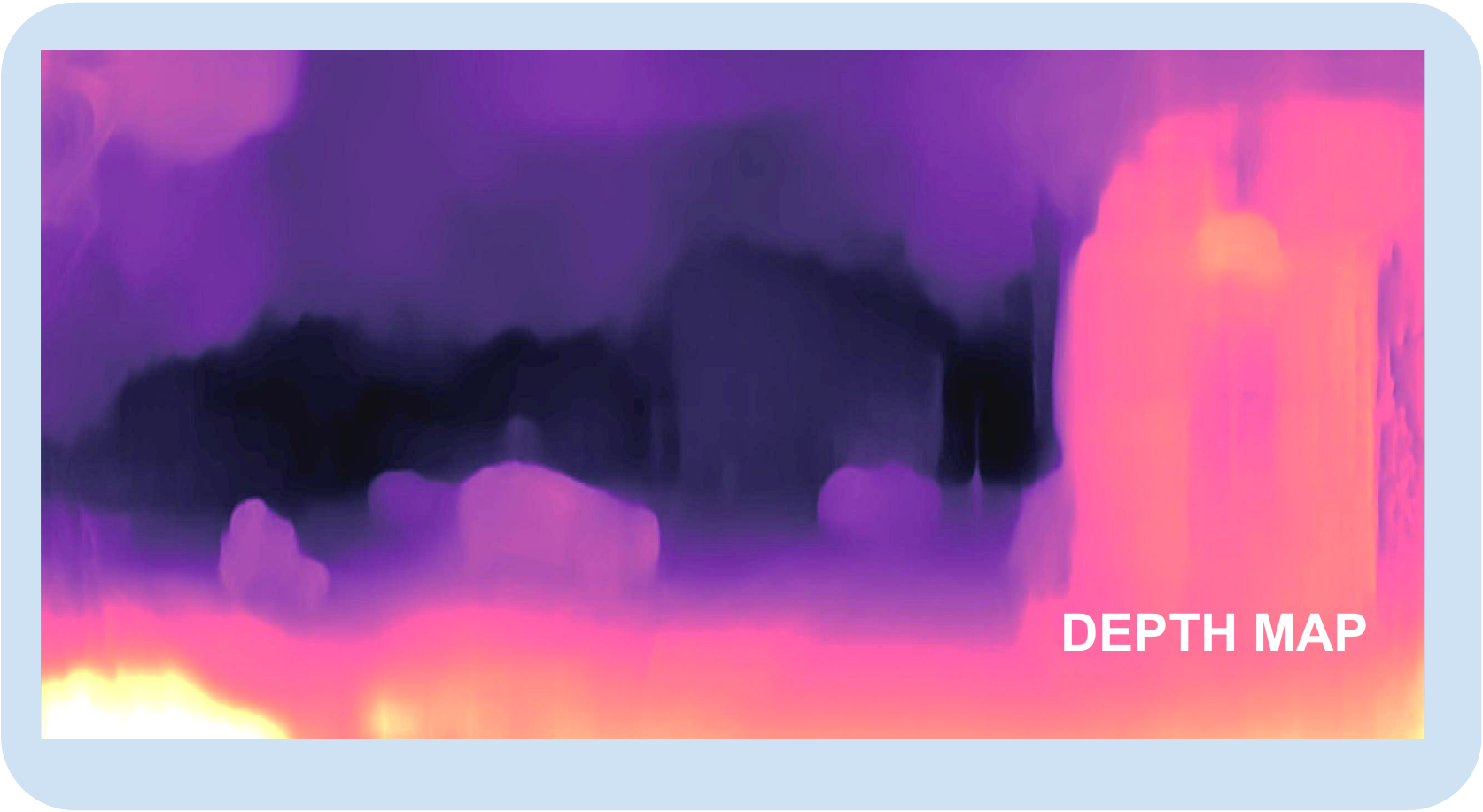} \\
 \hspace{-2mm} \text{(a) binaural sounds} & \hspace{-1mm} \text{(b) super-resolved ones} &
  \text{(c) semantic labels} & \hspace{-2mm} \text{(d) depth values} \\ 
  \caption{An illustration of our three tasks: (a) input binaural sounds, (b) super-resolved binaural sounds -- from azimuth angle $0^\circ$ to $90^\circ$, (c) auditory semantic perception for three sound-making object classes, and (d) auditory depth perception. }
  \end{tabular} \vspace{-4.5mm}
  \label{fig:teaser}
\end{figure} 


In particular, we propose a sensor setup containing eight professional binaural microphones and a $360^\circ$ camera. We used it to record the new `Omni Auditory Perception Dataset' on public streets. The semantic labels of the sound-making objects and the depth of the scene are inferred from the video frames using well-established vision systems as shown in Fig.~\ref{fig:teaser}. The co-existence of visual and audio cues is leveraged for supervision transfer to train the auditory perception systems without using human annotations. In particular, we employ a cross-modal framework that consists of vision teacher methods and sound student methods to let the students imitate the performance of the teachers. For evaluation, we have manually annotated a test set. The task of S$^{3}$R has accurate ground truth to train, thanks to our multi-microphone rig.
Finally, we formulate the semantic prediction task and the two auxiliary tasks into a multi-tasking network which can be trained in an end-to-end fashion.


We evaluate our method on our new Omni Auditory Perception Dataset.
Extensive experimental results reveal that 1) our method achieves good results for auditory semantic perception, auditory depth prediction and spatial sound super-resolution;  2) the three tasks are mutually beneficial -- training  them  together  achieves  the best results; and 3) both the number and orientations of microphones are important for auditory perception.


This work makes multiple contributions: 1) a novel approach to dense semantic label prediction for sound-making objects of multiple classes and an approach for dense depth prediction; 2) a method for spatial sound super-resolution, which is novel both as a standalone task and as an auxiliary task for auditory semantic prediction; and 3) a new Omni Auditory Perception Dataset with four pairs of binaural sounds ($360^\circ$ coverage) accompanied by synchronized $360^\circ$ videos which will be made publicly available.

\section{Related Works}
\label{sec:related}

\noindent
\textbf{Auditory Scene Analysis}. 
Sound segregation is a well-established research field aiming to organize sound into perceptually meaningful elements~\cite{auditory:scene:analysis:94}. Notable applications include background sounds suppression and speech recognition. Recent research has found that the motion from videos can be useful for the task \cite{separate:sound:watching:video:18,sounds:motion:iccv19}.  
Sound localization has been well studied with applications such as localizing sniper fire on the battle field, cataloging wildlife in rural areas, and localizing noise pollution sources in an urban environment. It also enriches human–robot interaction by complementing the robot’s perceptual capabilities~\cite{survey:sound:localization:15,localization:sound:source:review:17}. The task is often tackled by the beamforming technique with a microphone array~\cite{SoundCompass:2014}, with a notable exception that relies on a single microphone only~\cite{sound:location:09}. The recent advance of deep learning enables acoustic camera systems for real-time reconstruction of acoustic camera spherical maps~\cite{deepwave:nips19}. 

Auditory scene analysis has been widely applied to automotive applications as well. For instance, auditory cues are used to determine the
occurrence of abnormal events in driving scenarios~\cite{soundscape:icra17}. An acoustic safety emergency system has also been proposed~\cite{acoustic:based:safety:emergency:09}. Salamon et al. have presented a taxonomy of urban sounds and a new dataset, UrbanSound, for automatic urban sound classification~\cite{dataset:taxonomy:14}. The closest to our work is the very recent work of car detection with stereo sounds~\cite{vehicle:tracking:sound:iccv19} in which a 2D bounding-box is proposed for the sound-making object in an image frame. While being similar in spirit, our work differs significantly from theirs. Our method is designed for dense label prediction for multiple classes instead of a 2D bounding box prediction for a single class. Our method also includes dense depth prediction and spatial sound super-resolution. Binaural sounds are different from general stereo sounds and our method works with panoramic images to have omni-view perception. 
Another similar work to ours is ~\cite{irie2019seeing}. We differ significantly in: 1) having multiple semantic classes, 2) working in unconstrained real outdoor environment, and 3) a multi-task learning setup. 


\noindent
\textbf{Audio-Visual Learning}. 
There is a body of work to localize the sources of sounds in visual scenes~\cite{localize:sound:source:scene:cvpr18,separate:sound:watching:video:18,look:listen:learn:iccv17,objects:sound:eccv18,Harmony:in:Motion:cvpr07,sound:pixels:eccv18,cosegmenting:sounds:objects:iccv19,sounds:motion:iccv19}. The localization is mostly done by analyzing the consistency between visual motion cues and audio motion cues over a large collection of video data. These methods generally learn to locate image regions which produce sounds and separate the input sounds into a set of components that represent the sound for each pixel. Our work uses binaural sounds rather than monaural sounds. Our work localizes and recognizes sounds at the pixel level in image frames, but performs the task with sounds as the only inputs. 
Audio has also been used for estimating the geometric layout of indoor scenes~\cite{3d:room:geometry:audio-visual:sensors:17,glass:reconstruction:acoustic:15,room:geometry:acoustic:response:12,accoustic:echoes:reveal:room:shape:13}. The general idea is that the temporal relationships between the arrival times of echoes allows us to estimate the geometry of the environment. 
There are also works to add scene-aware spatial audio to 360$^{\circ}$ videos in typical indoor scenes~\cite{scene-aware-audio} by using a conventional mono-channel microphone and a speaker and by analyzing the structure of the scene.  The notable work by Owens et al.~\cite{Owens_2016_CVPR} have shown that the sounds of striking an object can be learned based on the visual appearance of the objects. By training a neural network to predict whether video frames and audio are temporally aligned~\cite{Owens_2018_ECCV}, audio representations can be learned. Our S$^3$R can be used as another self-learning method for audio representations.


\noindent
\textbf{Cross-domain Distillation}.
The interplay among senses is basic to the sensory organization in human brains \cite{merging:of:sense} and is the key
to understand the complex interaction of the physical world. Fortunately, most videos like those available in the Internet contain both visual information and audio information, which provide a bridge linking those two domains and enable many interesting learning approaches. Aytar et al. propose an audio-based scene recognition method by cross-domain distillation to transfer supervision from the visual domain to the audio domain~\cite{aytar2016soundnet}. A similar system was proposed for emotion recognition~\cite{emotion:cross-model:transfer:18}. Ambient sounds can provide supervision for visual learning as well~\cite{ambient:sound:eccv16}. 



\section{Approach}
\label{sec:approach}

\begin{figure*}[!tb]
  \centering
  \includegraphics[width=0.97\textwidth,height=0.35\textwidth]{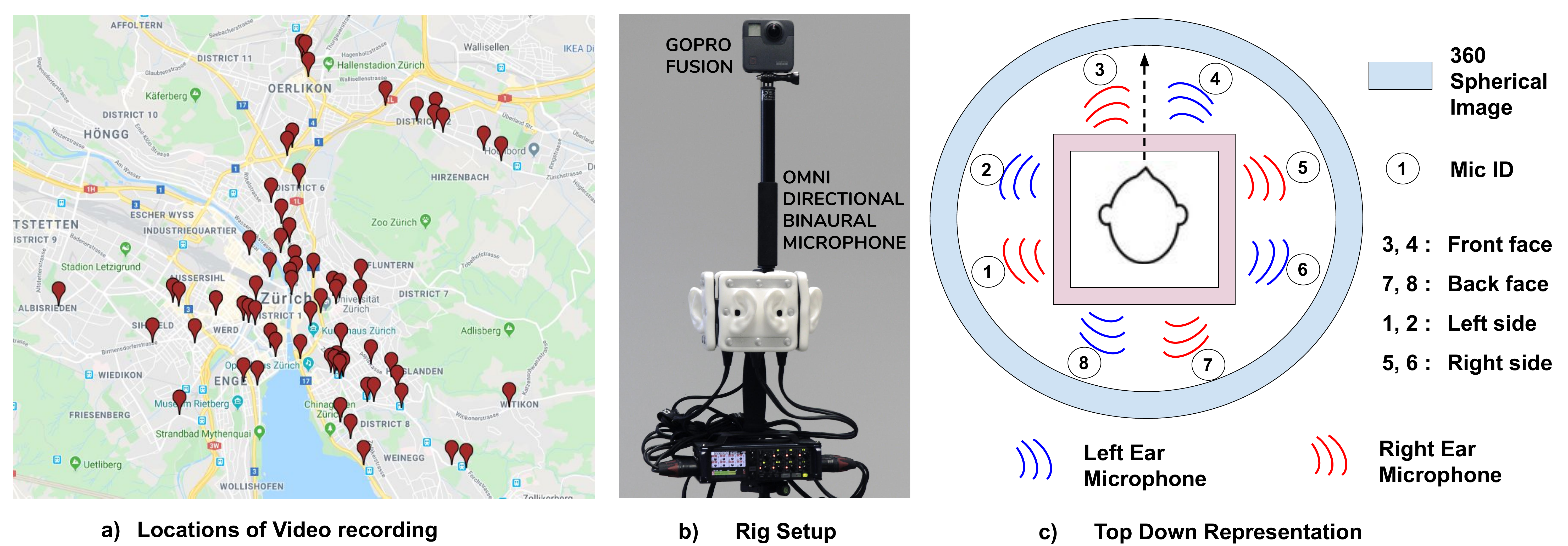}
  \caption{Sensor and dataset: a) data capture locations, b) our custom rig and c) abstract depiction of our recording setup with sensor orientations and microphone ids.}
  \label{fig:setup}\vspace{-1mm}
\end{figure*}


\vspace{-2mm}
\subsection{Omni Auditory Perception Dataset} 
\label{sec:dataset} 

Training our method requires a large collection of omni-directional binaural audios and accompanying $360^{\circ}$ videos.  Since no public video dataset fulfills this requirement, we collect a new dataset with a custom rig. As shown in Fig.~\ref{fig:setup}, we assembled a rig consisting of a 3Dio Omni Binaural Microphone, a $360^{\circ}$ GoPro Fusion camera and a Zoom F8 MultiTrack Field Recorder, all attached to a tripod. 
We mounted the GoPro camera on top of the 3Dio Omni Binaural Microphone by a telescopic pole to capture all of the sights with minimum occlusion from the devices underneath. This custom rig enables omni seeing and omni binaural hearing with 4 pairs of human ears. 
The 8 microphones are connected to 8 tracks of the MultiTrack Recorder. The recorder has 8 high-quality, super low-noise mic preamps to amplify the sounds, provides accurate control of microphone sensitivity and allows for accurate synchronization among all microphones. The GoPro Fusion camera captures 4K videos using a pair of $180^\circ$ cameras housed on the front and back faces. The two cameras perform synchronized image capture and are later fused together to render $360^\circ$ videos using the GoPro Fusion App. We further use the clap sound to synchronize the camera and 3Dio microphones. 
After the sensors are turned on, we do hand clapping near them. This clapping sounds recorded by both the binaural mics and the built-in mic of the camera are used to synchronize the binaural sounds and the video. The video and audio signals recorded by the camera are synchronized by default. Videos are recorded at $30$ fps. Audios are recorded at $96$ kHz.


We recorded videos on the streets of a big European City covering 165 locations within an area of 5km $\times$ 5km as shown in Fig.~\ref{fig:setup}(a). We choose the locations next to road junctions, where we kept the rig stationary for the data recording of the traffic scenes. For each location, we recorded data for around 5-7 minutes. 
Our dataset consists of 165 city traffic videos and audios with an average length of 6.49 minutes, totalling 15 hours. We post-process the raw video-audio data into 2 second segments, resulting in $64,250$ video clips. The videos contain numerous sound-making objects such as cars, trams, motorcycles, pedestrians, buses and trucks. 

It is worth noticing that the professional 3Dio binaural mics simulate how human ears receive sound, which is different from the general stereo mics or monaural mics. Humans localize sound sources by using three primary cues \cite{localization:sound:source:review:17}: interaural time difference (ITD), interaural level difference (ILD), and head-related transfer function (HRTF). ITD is caused by the difference between the times sounds reach the two ears. ILD is caused by the difference in sound pressure level reaching the two ears due to the acoustic shadow casted by the listener's head. HRTF is caused because the pinna and head affect the intensities of sound frequencies. 
All these cues are missing in monaural audio, thus in this work, we focus on learning to localize semantic objects with binaural audios. This difference makes our dataset more suitable for cognitive applications.

\begin{figure*}[!tb]
  \centering
  \includegraphics[width=1.0\textwidth]{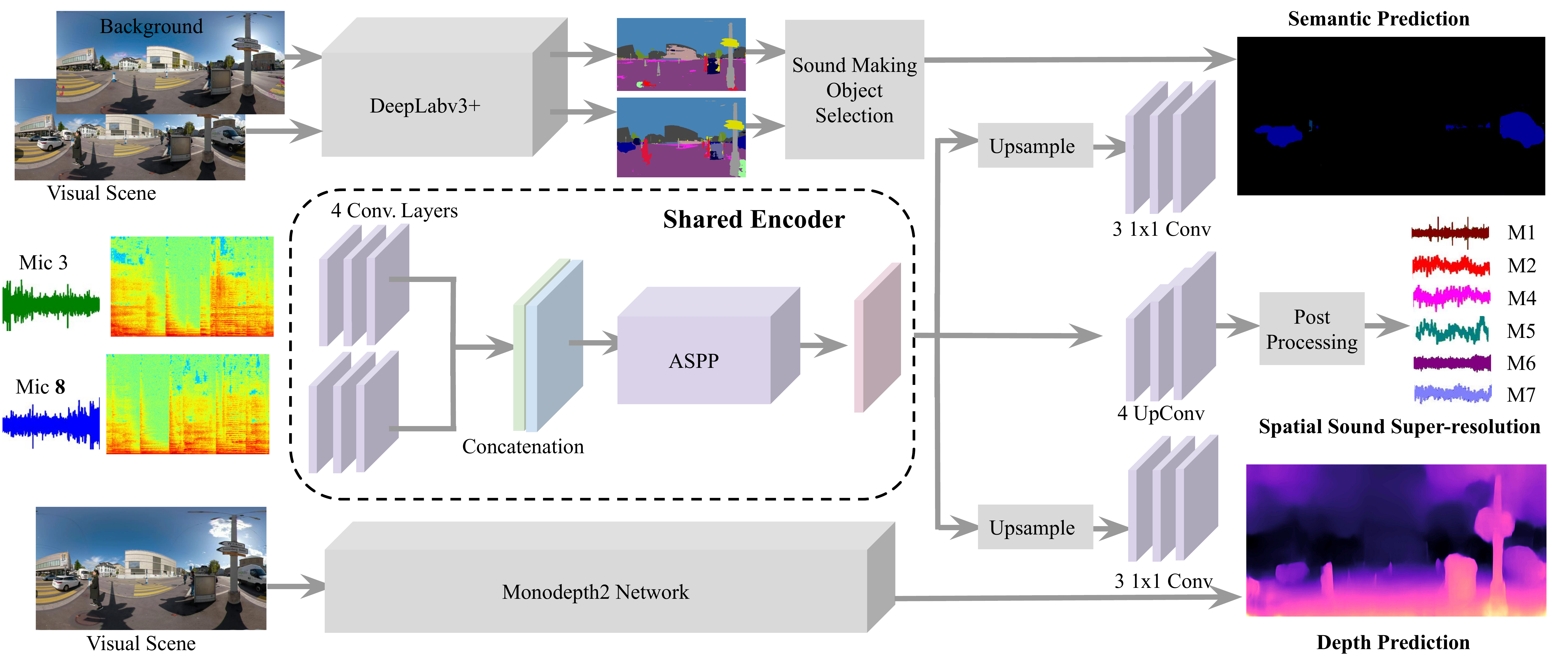}
  \caption{The diagram of our method for the three considered tasks. The encoder is shared by all tasks and each task has its own decoder.}
  \label{fig:pipeline}\vspace{-2mm}
\end{figure*}

\vspace{-2mm}
\subsection{Auditory Semantic Prediction} 
\label{sec:semantic:obj:loc}
Since it is costly to create a large collection of human annotations, we follow a teacher-student learning strategy to transfer knowledge from vision to audio~\cite{aytar2016soundnet,vehicle:tracking:sound:iccv19}. Thus, our auditory semantic object prediction system is composed of two key components: a teacher vision network and a student audio network. The difference to previous methods is that we learn to transfer precise semantic segmentation results instead of scene labels \cite{aytar2016soundnet} or bounding boxes \cite{vehicle:tracking:sound:iccv19}.

\noindent
\textbf{Vision network}. We employ the DeepLabv3+ model \cite{chen2018encoder} as it is the current state-of-the-art (s-o-t-a) semantic segmentation model. We pick the middle frame of our 2-second video clip as the target frame and feed it to the teacher network to generate the semantic map. During training, each target frame is fed into a Cityscapes \cite{Cordts2016Cityscapes} pre-trained DeepLabv3+ to assign a semantic label to each pixel. Since objects in many classes such as \emph{sky}, \emph{road} and \emph{parked cars} are not sound-making, it is very challenging to predict their semantic masks. Therefore, an object selection policy needs to be designed in order to collect the semantic masks of major sound-making objects. 

\noindent
\textbf{Sound-making Object Collection}.
Our dataset contains numerous sound making objects such as cars, trams, motorcycles, pedestrians, buses and bicycles. The target objects must be constrained to make sure that the task is challenging but still achievable by current sensing systems and learning methods. In this work, we focus on \emph{car}, \emph{tram}, \emph{motorcycle} due to their high occurrences in the datasets and because of them producing sufficient noise, as long as they move. 

As to the motion status, we employ background subtraction to remove the background classes, such as road, building and sky, and the stationary foreground classes such as parked cars. In the end, only the semantic masks of moving trams, moving cars and moving motorcycles are taken as the prediction targets. This selection guides the attention of the learning method to major sound-making objects and avoids localizing `rare' sound-making objects and sound-irrelevant objects such as \emph{parked car} and \emph{sky}.  

There is a rich body of studies for background estimation~\cite{background:substraction:11}, in this work we employ a simple method based on majority voting. The method works surprisingly well. Specifically, the background image is computed as 
\begin{equation}
I_{bg}(h,w) = \text{Mode}\{I_{1}(h,w), I_{2}(h,w), ..., I_T(h,w) \},
\label{eq:modeOp} 
\end{equation}
where $T$ is the length of the complete video sequence, $(h,w)$ are pixel indexes, and $\text{Mode}\{.\}$ computes the number which appears most often in a set of numbers.  Since the complete video sequence is quite long (about 5-7 mins), the background estimation is accurate and reliable. 

The sound-making objects are detected by the following procedure: given an video frame $I_t$ and its corresponding background image $I_{bg}$, we use DeepLabv3+ to get their semantic segmentation results $Y_t$, and $Y_{bg}$. Fig.~\ref{fig:pipeline} gives an illustration. The detection is done as
\begin{equation}
    \label{eq:sounding:obj} 
    S(h,w) = \left\{ 
    \begin{array}{rl}
    1 & \text{if  } Y_t(h,w) \in \{car, train, motorcycle\} \\ 
    & \text{ and } Y_t(h,w) \neq Y_{bg}(h,w),\\
    0 & \text{otherwise},
   \end{array} 
   \right.
\end{equation}
where $1$ indicates pixel locations of sound-making objects and $0$ otherwise.  
Fig.~\ref{fig:result1} shows examples of the detected background and the detected sound-making target objects (i.e. ground truth). 

\noindent
\textbf{Audio network}.
We treat auditory semantic prediction from the binaural sounds as a dense label prediction task. We take the semantic labels produced by the teacher vision network and filter by Eq.~\ref{eq:sounding:obj} as 
pseudo-labels, and then train a student audio network (BinauralSematicNet) to predict the pseudo semantic labels directly from the audio signals. A cross-entropy loss is used. The description of the network architecture can be found in Sec.~\ref{sec:network:architecture} and in the supple. material.  

\vspace{-2mm}
\subsection{Auditory Depth Perception} 
\label{sec:depth:prediction} 
Similar to auditory semantic object localization, our auditory depth prediction method is composed of a teacher vision network and a student audio network. It provides auxiliary supervision for semantic perception task.

\noindent
\textbf{Vision network}. We employ the MonoDepth2 model \cite{monodepth2:iccv19} given its good performance. We again pick the middle frame of our 2-second video clip as the target frame and feed it to the teacher network to generate the depth map. The model is pre-trained on KITTI~\cite{Geiger2013IJRR}. We estimate the depth for the whole scene, similar to previous methods for holistic room layout estimation with sounds \cite{room:geometry:acoustic:response:12,3d:room:geometry:audio-visual:sensors:17}. 


\noindent
\textbf{Audio network}.
We treat depth prediction from the binaural sounds as a dense regression task. We take the depth values produced by the teacher vision network as 
pseudo-labels, and then train a student audio network
(BinauralDepthNet) to regress the pseudo depth labels directly from the audio signals. The L2 loss is used. The description of the network architecture can be found in Sec.~\ref{sec:network:architecture}.  

\vspace{-2mm}
\subsection{Spatial Sound Super-resolution (S$^3$R)}
\label{sec:sasr} 
We leverage our omni-directional binaural microphones to design a novel task of spatial sound super-resolution (S$^3$R), to provide auxiliary supervision for semantic perception task.
The S$^3$R task is motivated by the well-established studies~\cite{sound:localization:head:movement:14,role:head:movements:sound:localization:40} about the effects of head movement in improving the accuracy of sound localization. 
Previous studies also found that rotational movements of the head occur most frequently during source localization~\cite{sound:localization:head:movement:67}.
Our omni-directional binaural microphone set is an ideal device to simulate head rotations at different discrete angles.  
Inspired by these findings, we study the effect of head rotation on auditory semantic and depth perception. It is very challenging to make a rotating sensor rig during signal presentation. Hence, we leverage the fact that our device has four pairs of binaural microphones resembling sparsely sampled head rotations, i.e. to an azimuth angle of $0^{\circ}$ $90^{\circ}$, $180^{\circ}$ and $270^{\circ}$, respectively.  


Specifically, the S$^3$R task is to train a neural network to predict the binaural audio signals at other azimuth angles given the signals at the azimuth angle of $0^\circ$.  
We denote the received signal by the left and right ears at azimuth $0^{\circ}$ by $x^{L_0}(t)$ and $x^{R_0}(t)$, respectively. We then feed those two signals into a deep network to predict the binaural audio signals $x^{L_\alpha}(t)$ and $x^{R_\alpha}(t)$ at azimuth $\alpha^{\circ}$. Inspired by~\cite{gao20192}, we predict the difference of the target signals to the input signals, instead of directly predicting the absolute values of the targets. This way, the network is forced to learn the subtle difference. Specifically, we predict the difference signals:
\begin{equation}
\begin{split}
   & x^{DL_{\alpha}}(t) =  x^{L_0}(t) -  x^{L_\alpha}(t)   \\
   & x^{DR_{\alpha}}(t) =  x^{R_0}(t) -  x^{R_\alpha}(t), 
   \end{split}
\end{equation}
where $\alpha \in \{ 90^{\circ}, 180^{\circ}, 270^{\circ}\}$. In order to leverage the image processing power of convolutional neural network, we follow the literature and choose to work with the spectrogram representation. Following ~\cite{gao20192}, real and imaginary components of complex masks are predicted. The masks are multiplied with input spectrograms to get the spectrograms of the difference signals; the raw waveforms of the difference signals are then produced by applying Inverse Short-time Fourier Transform (ISTFT) \cite{invert:fourier:transform:83}; and finally the target signals are reconstructed by adding back the reference raw waveform.

\vspace{-2mm}
\subsection{Network Architecture} 
\label{sec:network:architecture} 

Here, we present our multi-tasking audio network for all the three tasks. The network is composed of one shared encoder and three task-specific decoders.
The pipeline of the method is shown in Fig.~\ref{fig:pipeline}. As to the encoder, we convert the two channels of binaural sounds to log-spectrogram representations. Each spectrogram is passed through 4 strided convolutional (conv) layers with shared weights before they are concatenated. Each conv layer performs a $4\times4$ convolution with a stride of $2$. Each conv layer is followed by a BN layer and a ReLU activation. The concatenated feature map is further passed to a Atrous Spatial Pyramid Pooling (ASPP) module~\cite{chen2017deeplab}. ASPP has one $1\times1$ convolution and three $3\times3$ convolutions with dilation rates of $6$, $12$, and $18$. 
Each of the convolutions has $64$ filters and a BN layer. 
ASPP concatenates all the features and passes them through a $1\times1$ conv layer to generate binaural sound features. This feature map is taken as the input to our decoders.

Below, we present the three task-specific decoders.
For the semantic prediction task, we employ a decoder to predict the dense semantic labels from the above feature map given by the shared encoder. The decoder comprises of an upsampling layer and three $1\times1$ conv layers. For the first two conv layers, each is followed by a BN and a ReLU activation; for the last one, it is followed by a softmax activation. We use the same decoder architecture for the depth prediction task, except that we use ReLU activation for the final conv layer of the decoder. For the S$^3$R task, we perform a series of $5$ up-convolutions for the binaural feature map, each convolution layer is followed by a BN and a ReLU activation. The last layer is followed by a sigmoid layer which predicts a complex valued mask. We perform a few post processing steps to convert this mask to binaural sounds at other azimuth angles as mentioned in Sec.~\ref{sec:sasr}.


\smallskip
\noindent
\textbf{Loss function.} We train the complete model shown in Fig.~\ref{fig:pipeline} in an end-to-end fashion. We use a) cross-entropy loss for the semantic prediction task which is formulated as dense pixel labelling to $3$ classes, b) L2 loss for the depth prediction task to minimize the distance between the predicted depth values and the ground-truth depth values, and c) L2 loss for the S$^{3}$R task to minimize the distance between the predicted complex spectrogram and the ground truths. Hence, the total loss $L$ for our multi-tasking learning is 
\begin{equation}
   L = L_{semantic} + \lambda_{1}L_{depth} + \lambda_{2}L_{s^3r}\\
   \label{eqn:loss}
\end{equation}
where $\lambda_{1}$ and $\lambda_{2}$ are weights to balance the losses. 
The detailed network architecture will be provided in the supple. material. 



\section{Experiments}
\label{sec:expts}


\noindent
\textbf{Data Preparation}.
Our dataset comprises of $64,250$ video segments, each of 2 seconds long. We split the samples into three parts: $51,400$ for training, $6,208$ for validation and $6,492$ for testing. We use 2-seconds segments following ~\cite{vehicle:tracking:sound:iccv19}, which shows that performances are stable for $\geq$1 second segments. For each scene, a background image is also precomputed according to Eqn.~\ref{eq:modeOp}. For the middle frame of each segment, we generate the ground truth for semantic segmentation task by using the Deeplabv3+~\cite{chen2018encoder} pretrained on Cityscapes dataset~\cite{Cordts2016Cityscapes} and the depth map by using  the Monodepth2~\cite{monodepth2:iccv19} pretrained on KITTI dataset~\cite{Geiger2013IJRR}.  
We use \emph{AuditoryTestPseudo} to refer the test set generated this way. In order to more reliably evaluate the method, we manually annotate the middle frame of $80$ test video segments for the three considered classes, namely car, train and motorcycle. We carefully select the video segments such that they cover diverse scenarios such as daylight, night, foggy and rainy.
We use LabelMeToolbox~\cite{russell2008labelme} for the annotation and follow the annotation procedure of Cityscapes~\cite{Cordts2016Cityscapes}. We call this test set \emph{AuditoryTestManual}.

For all the experiments, a training or a testing sample consists of a 2-second video segment and eight 2-second audio channels. We preprocess audio samples following techniques from ~\cite{sound:pixels:eccv18,gao20192}. We keep the audio samples at 96kHz and their amplitude is normalized to a desired RMS level, which we set as $0.1$ for all the audio channels. For normalization, we compute mean RMS values of amplitude over the entire dataset separately for each channel. An STFT is applied to the normalized waveform, with a window size of $512$ ($5.3$ms), hop length of $160$ ($1.6$ms) resulting a Time-Frequency representation of size of $257\times601$ pixels. Video frames are resized to $960\times1920$ pixels to fit to the GPU.

\smallskip
\noindent
\textbf{Implementation Details}. We train our complete model using Adam solver~\cite{kingma2014adam} with a learning rate of 0.00001 and we set a batch size of $2$. We train our models on GeForce GTX 1080 Ti GPUs for 20 epochs. For joint training of all the three tasks, we keep $\lambda_{1}=0.2$ and $\lambda_{2}=0.2$ in Eq.~\ref{eqn:loss}. 


\smallskip
\noindent
\textbf{Evaluation metrics}. We use the standard mean IoU for the semantic prediction task.
For audio super resolution, we use MSE error for the  spectrograms and the envelope error for the waveforms as used  in~\cite{gao20192}. For depth prediction, we employ RMSE, MSE, Abs Rel and Sq Rel by following~\cite{monodepth2:iccv19}.

\begin{table}[!tb]
  \centering
  \setlength\tabcolsep{2.0pt}
  \begin{adjustbox}{max width=\textwidth,max totalheight=\textheight}
  \begin{tabular}{lcccccccccccccc}
\toprule
Methods & \multicolumn{2} {c} {Microphone} & \multicolumn{2} {c} {Auxiliary Tasks} & \multicolumn{4} {c} {AuditoryTestPseudo} & \multicolumn{4} {c} {AuditoryTestManual} \\
 &  Mono & Binaural & S$^3$R & Depth & Car & MC & Train & All & Car & MC & Train & All \\ \midrule
BG  & & & & & 8.79 & 4.17 & 24.33 & 12.61 & - & - & - & - \\
Visual  & & & & & - & - & - & - & 79.01 & 39.07 & 77.34 & 65.35 \\ \cdashline{1-13}
Mono   &  \cmark & & & & 33.53 & 7.86 & 24.99 & 22.12 & 30.13 & 9.21 & 24.1 & 21.14 \\ 
Ours(B)   &   & \cmark & & & 35.80 & 19.51 & 40.71 & 32.01 & 35.30 & 13.28& 35.48& 28.02\\
Ours(B:D)  &   & \cmark &   &  \cmark & 33.53 & 28.01 &  55.32 & 38.95  & 32.42 & 25.8 & 50.12 & 36.11 \\ 
Ours(B:S)  &   & \cmark &  \cmark & & 35.62 & 36.81 & \textbf{56.49}  & 42.64 & \textbf{38.12} & 26.5 & 49.02 & 37.80 \\ 
Ours(B:SD)  &   & \cmark &  \cmark &  \cmark & \textbf{35.81} & \textbf{38.14} & 56.25  & \textbf{43.40}  & 35.51 & \textbf{28.51} & \textbf{50.32} & \textbf{38.01} \\ 
\bottomrule 
\end{tabular} 
\end{adjustbox}
\caption{Results of auditory semantic prediction. The results of DeepLabv3+ on the background image (BG) and on the target middle frame (Visual) are reported for reference purpose.  mIoU (\%) is used. MC denotes Motorcycle.} 
  \label{tab:maintable}
\end{table} \vspace{-2mm}

\vspace{-2mm}
\subsection{Auditory Semantic Prediction}

We compare the performance of different methods and report the results in Tab.~\ref{tab:maintable}. The table shows our method learning with sounds can generate promising results for  dense semantic object prediction.  We also find that using binaural sounds \emph{Ours(B)} generates significant better results than using \emph{Mono} sound. This is mainly because the major cues for sound localization such as ILD, ITD, and HRTF are missing in \emph{Mono} sound. 
Another very exciting observation is the joint training with the depth prediction task and the S$^3$R task are beneficial to the semantic prediction task. We speculate that this is because all the three tasks benefit from a same common goal -- reconstructing 3D surround sounds from Binaural sounds. Below, we present our ablation studies.     



\smallskip
\noindent 
\textbf{S$^{3}$R and depth prediction both give a boost}. As can be seen in Tab.~\ref{tab:maintable}, by adding S$^{3}$R or depth prediction as an auxiliary task, indicated by \emph{Ours(B:S)} and \emph{Ours(B:D)} respectively, improves the performance of our baseline \emph{Ours(B)} clearly. 
We also observe that using both of the auxiliary tasks together, indicated by \emph{Ours(B:SD)}, yields the best performance. This shows that both S$^3$R and depth helps. This is because the three tasks share the same goal -- extracting spatial information from binaural sounds.

\begin{figure*}[!tb]
\begin{tabular}{cccc}
  \centering \hspace{-2mm}
  \includegraphics[trim=160 20 140 65,clip,width=0.38\textwidth,height=0.2\textwidth]{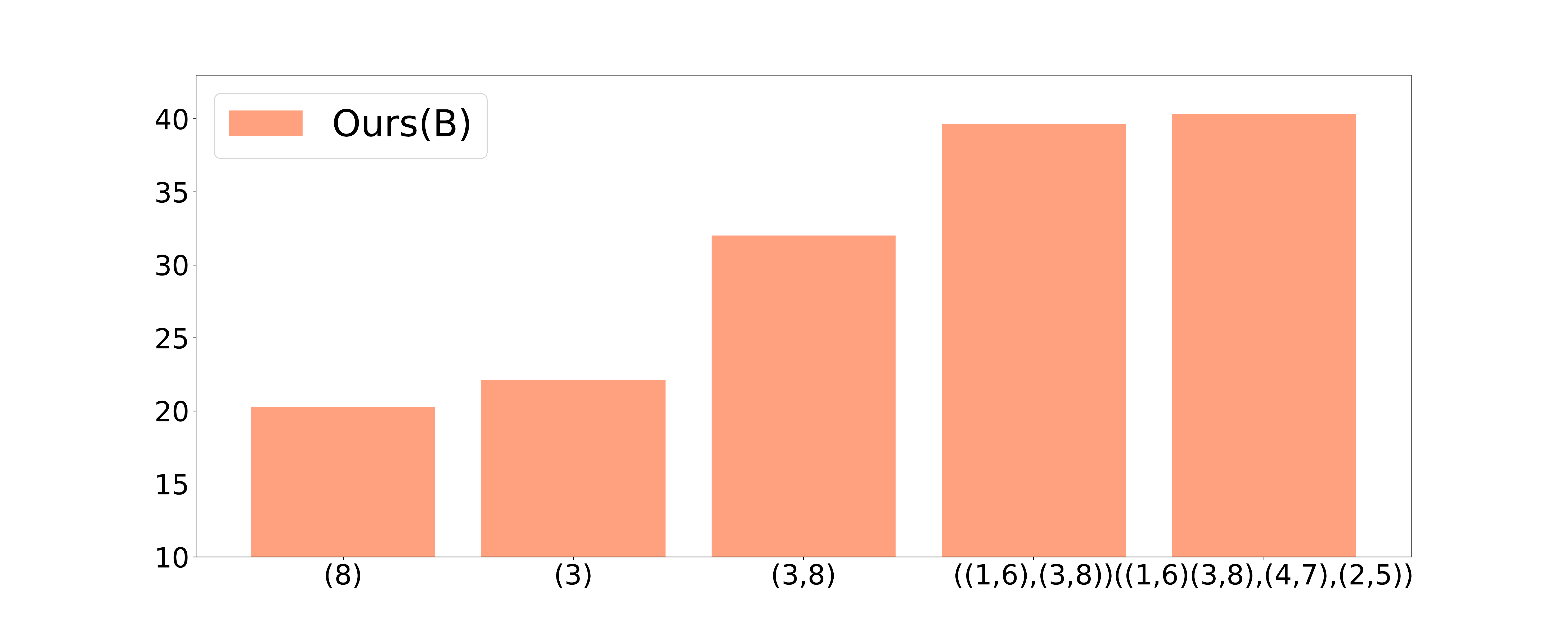} & \hspace{-2mm}
  \includegraphics[trim=45 10 50 45,clip,width=0.22\textwidth,height=0.2\textwidth]{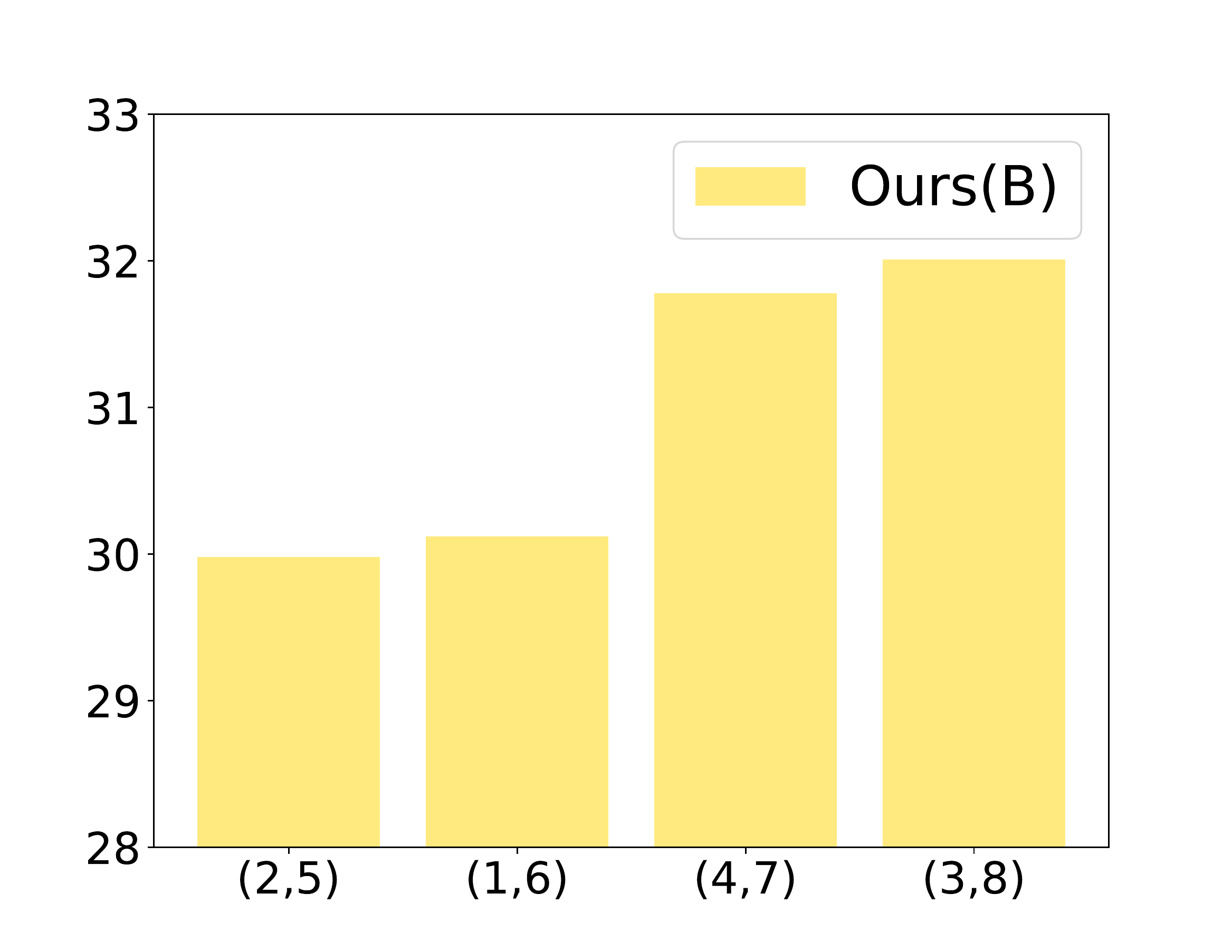} & \hspace{-2mm}
  \includegraphics[trim=150 20 100 65,clip,width=0.42\textwidth,height=0.2\textwidth]{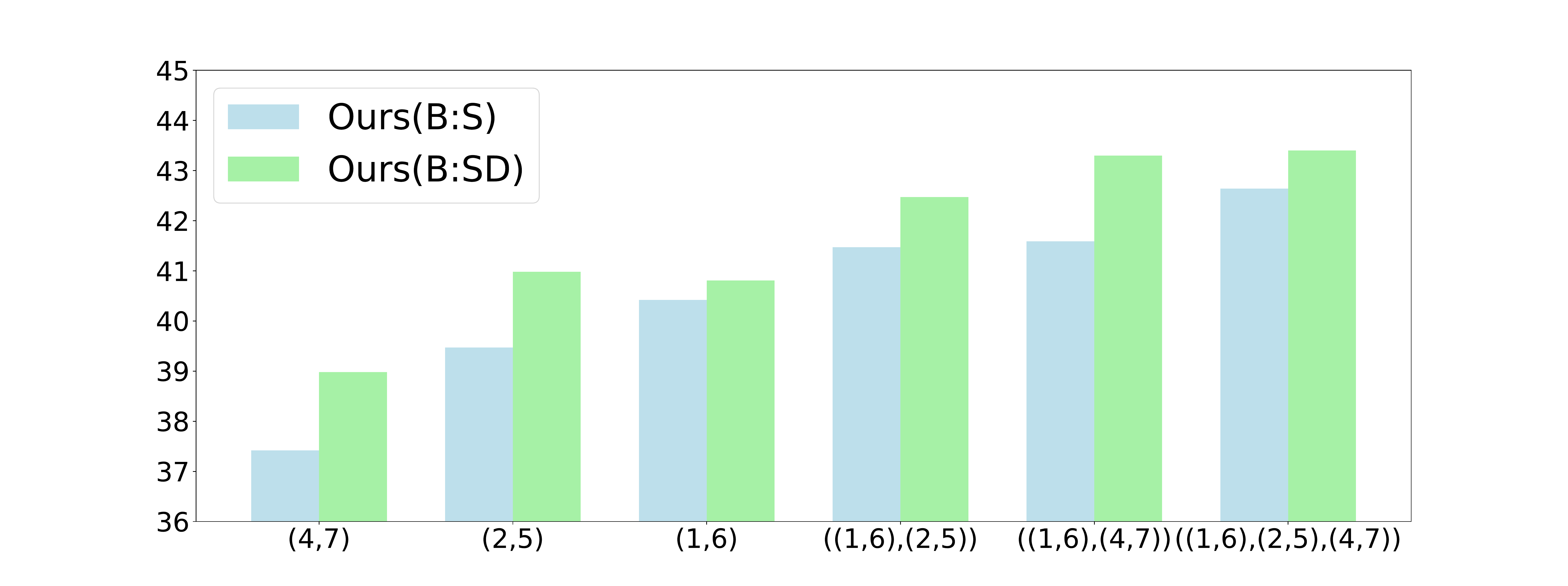}  
  \\
  \text{(a) \#(input mics)}  &
  \text{(b) input mic pairs} &\text{(c) Combination of output mics} 
  \\ 
  \caption{Semantic prediction ablation study results (mIoU) with different set of microphone used as inputs under \emph{Ours(B)} in (a) and (b) and ablation on output microphones for S$^3$R in \emph{Ours(B:S)} and \emph{Ours(B:SD)} in (c).}
  \label{fig:mic_ablation}
  \end{tabular} \vspace{-6.5mm}
\end{figure*}

\smallskip
\noindent
\textbf{Adding input channels increases the performance}. We compare the auditory semantic prediction (without using auxiliary tasks) accuracies in Fig.~\ref{fig:mic_ablation}(a) for different set of input microphones. Here, we experiment with a) Mono sound from 3 (front) or 8 (back) microphone, b) binaural sounds from the pair (3,8), c) 2 pairs of binaural sound channels ((1,6),(3,8)) which faces in four orthogonal directions, and d) 4 pairs of binaural sound channels. We see that semantic prediction accuracy increases from $22.12$\% when using \emph{Mono} sound to $40.32$\% when using all 8 channels. This shows that semantic prediction improves with the access to more sound channels.
 
\smallskip
\noindent
\textbf{Orientation of microphones matter}. 
Fig.~\ref{fig:mic_ablation}(b) shows the auditory semantic prediction results (without using auxiliary tasks) from different orientations of the input binaural pairs for the same scene. The pair (3,8) is aligned in parallel with the front facing direction 
of the camera as can be seen in Fig.~\ref{fig:setup}. We define this as orientation of 0\degree. 
Then, we have other pairs (1,6), (4,7) and (2,5) orientating at azimuth angles of 90\degree, 180\degree and 270\degree respectively. 
We observe that (3,8) outperforms all other pairs. 

\begin{figure*}[tb]
    \centering
    \includegraphics[width=0.24\textwidth]{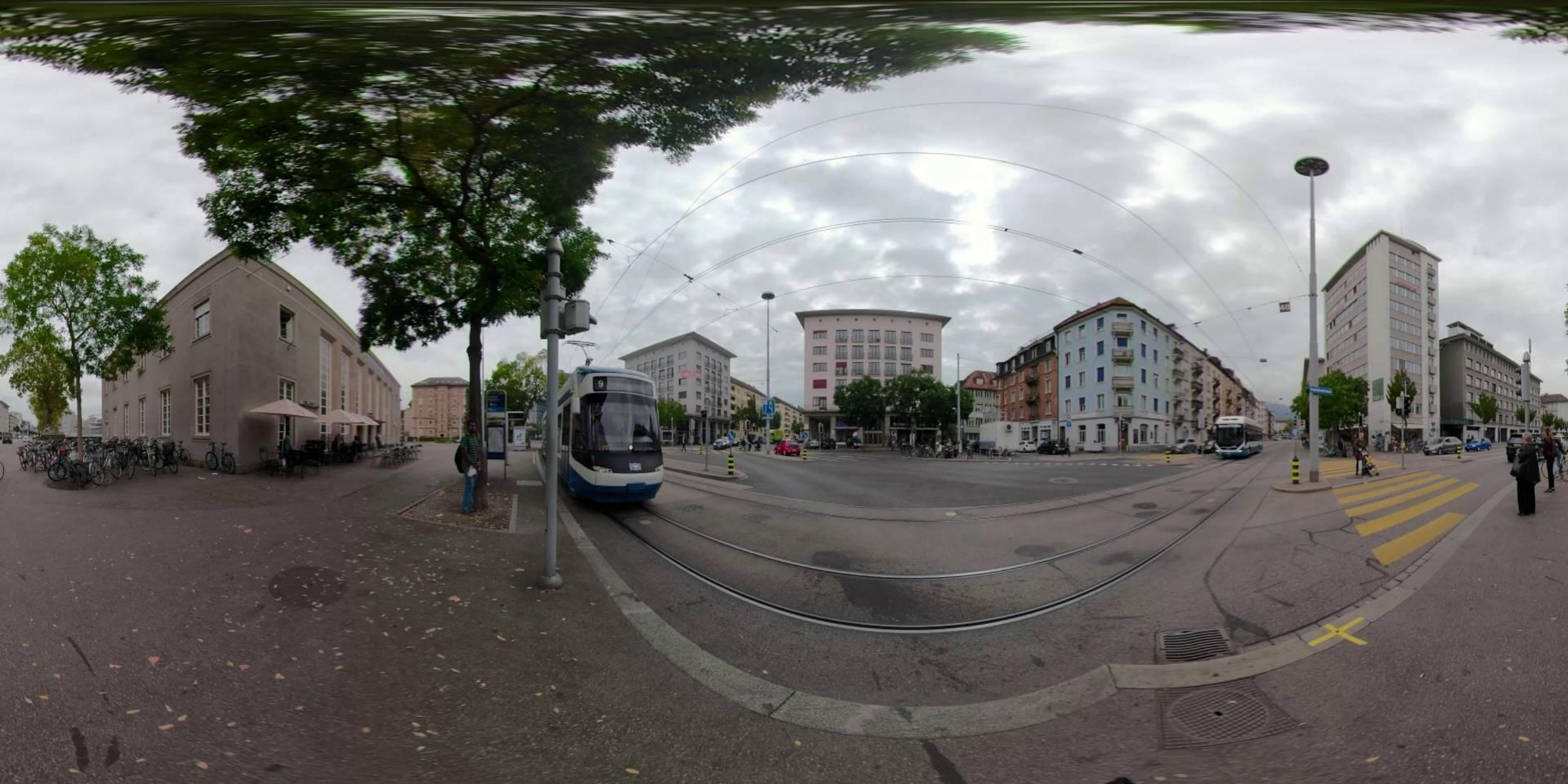}
    \hfil
    \includegraphics[width=0.24\textwidth]{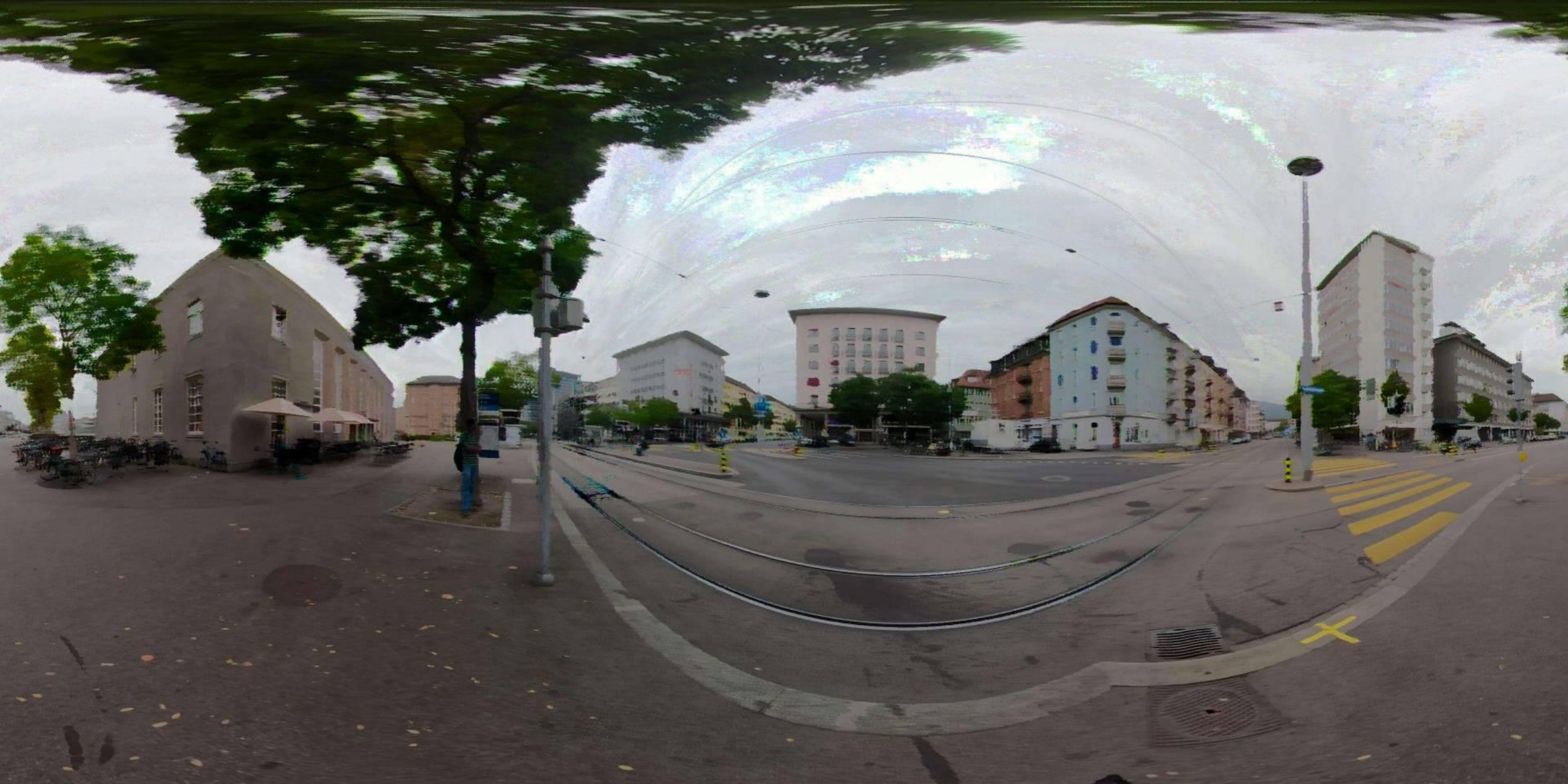}
    \hfil
    \includegraphics[width=0.24\textwidth]{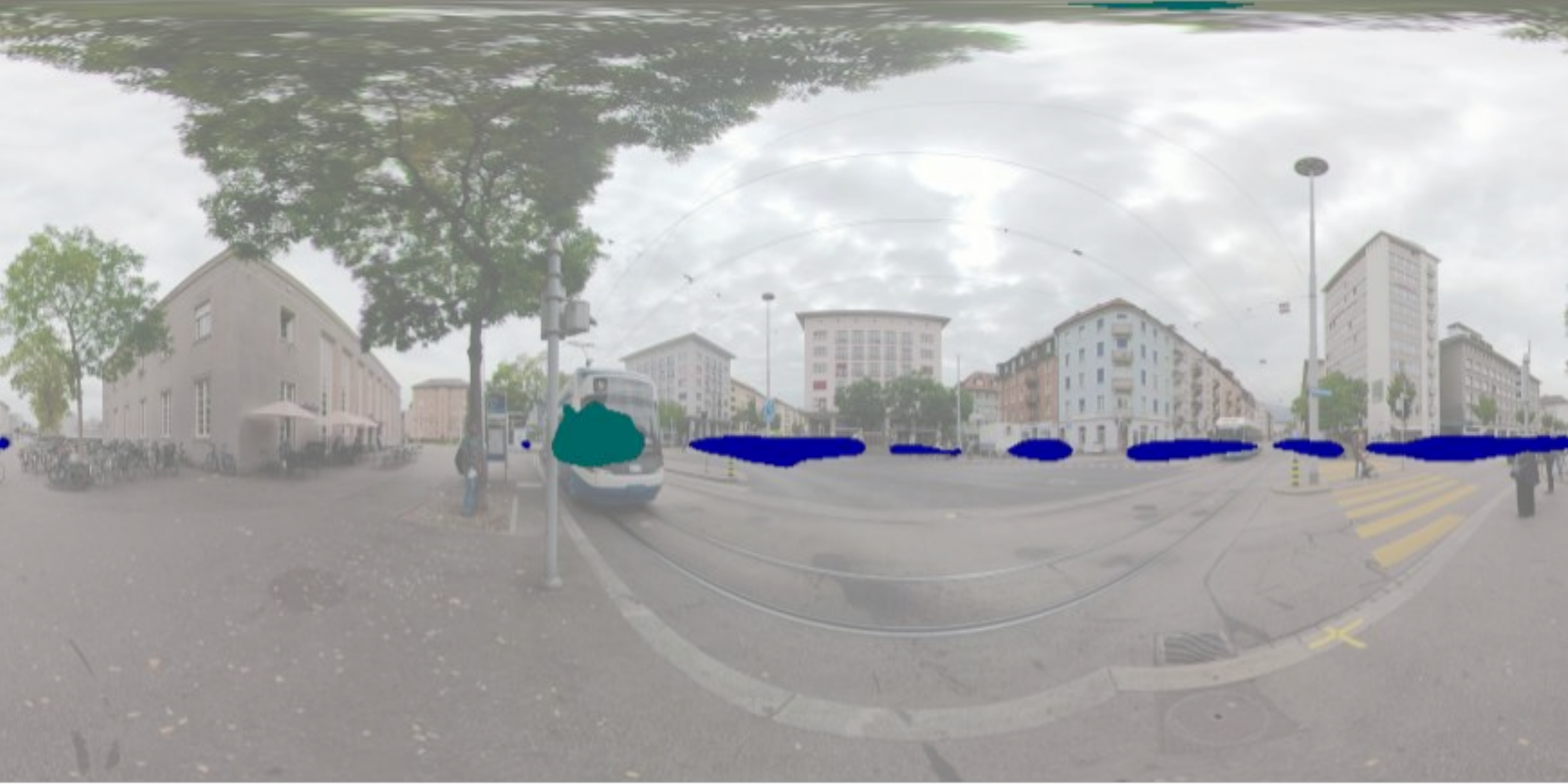}
    \hfil
    \includegraphics[width=0.24\textwidth]{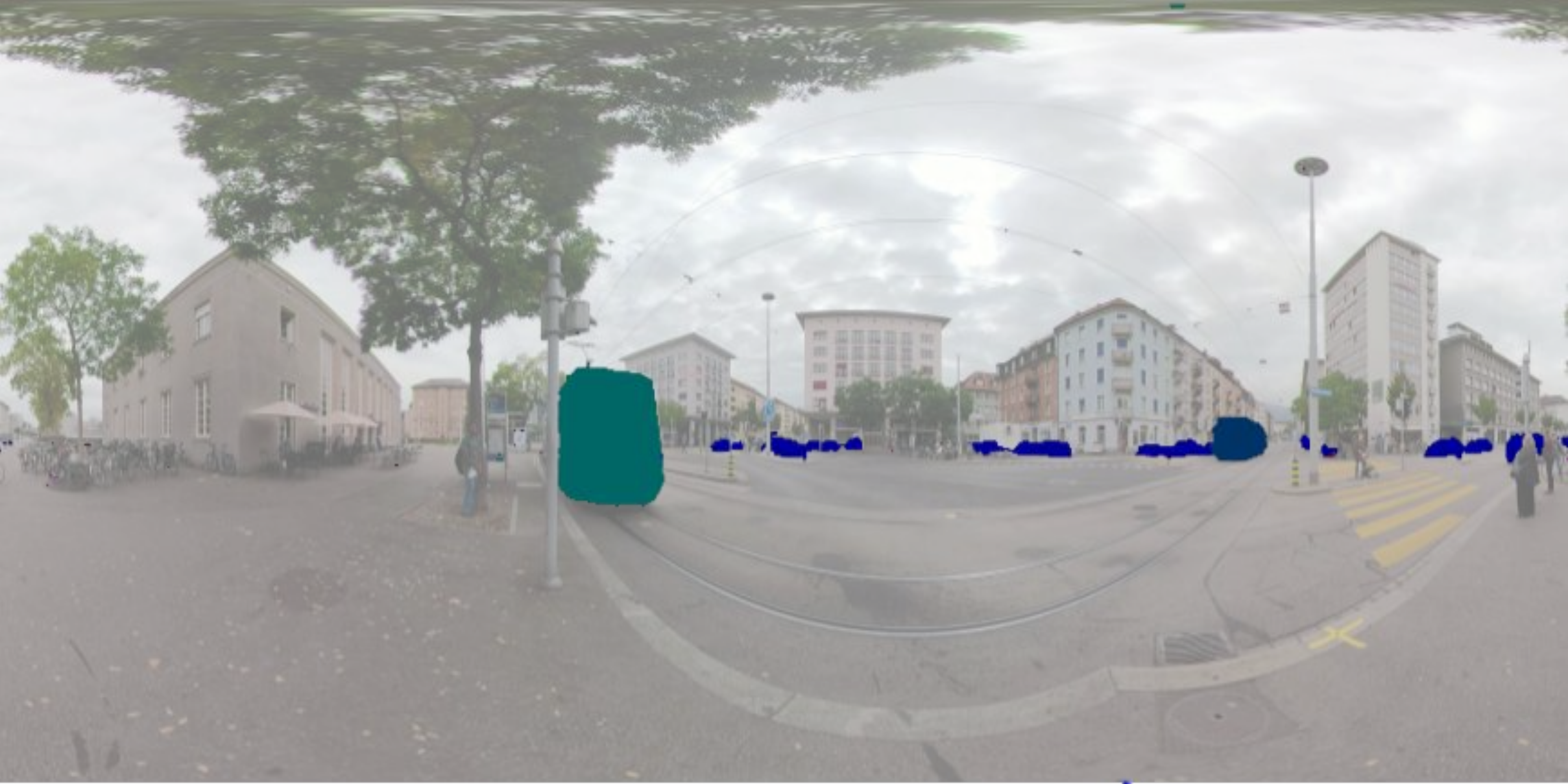}
    \\ \vspace{1mm}
    \includegraphics[trim=0 40 0 0,clip,width=0.24\textwidth,height=0.12\textwidth]{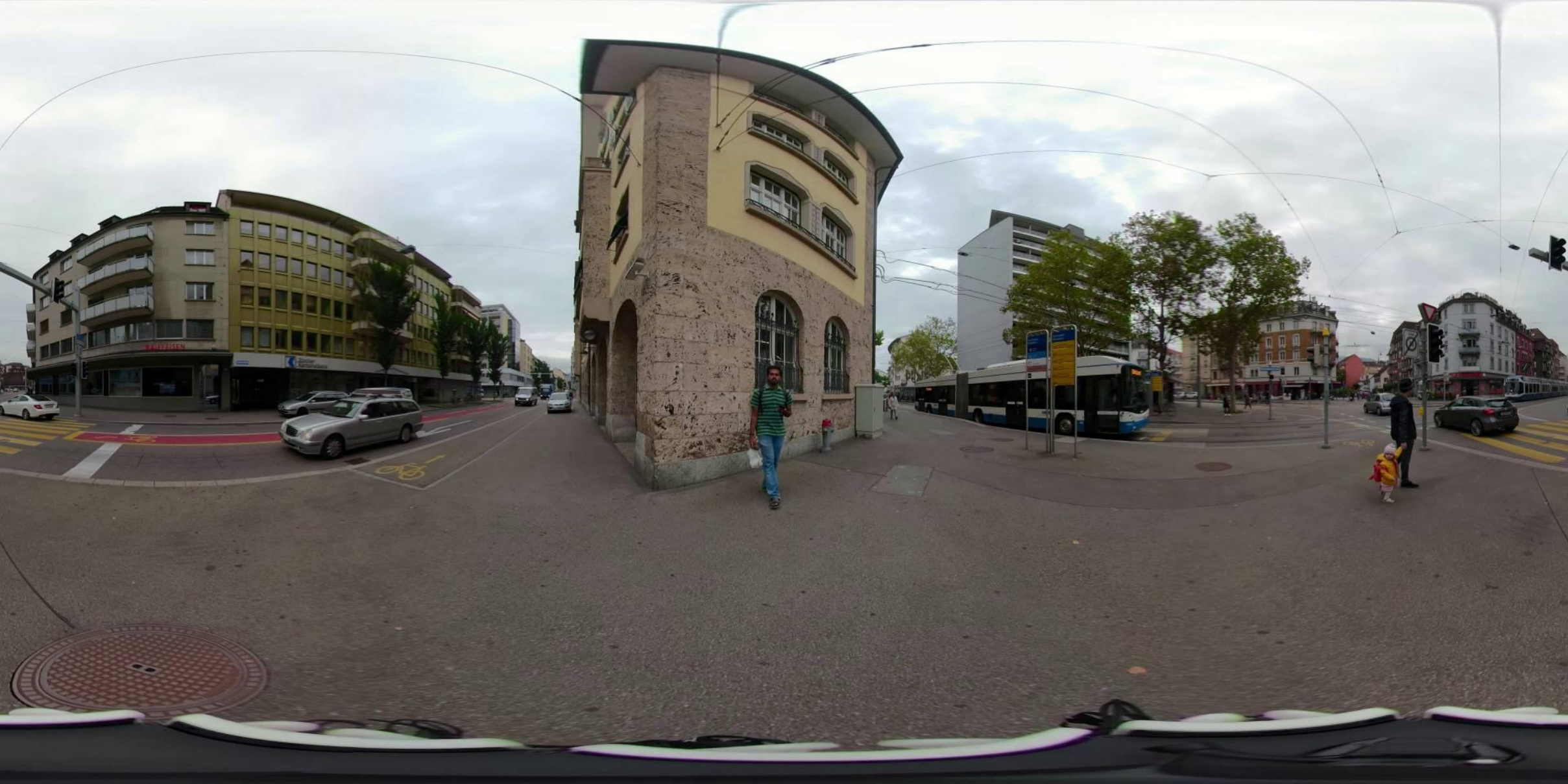}
    \hfil
    \includegraphics[width=0.24\textwidth]{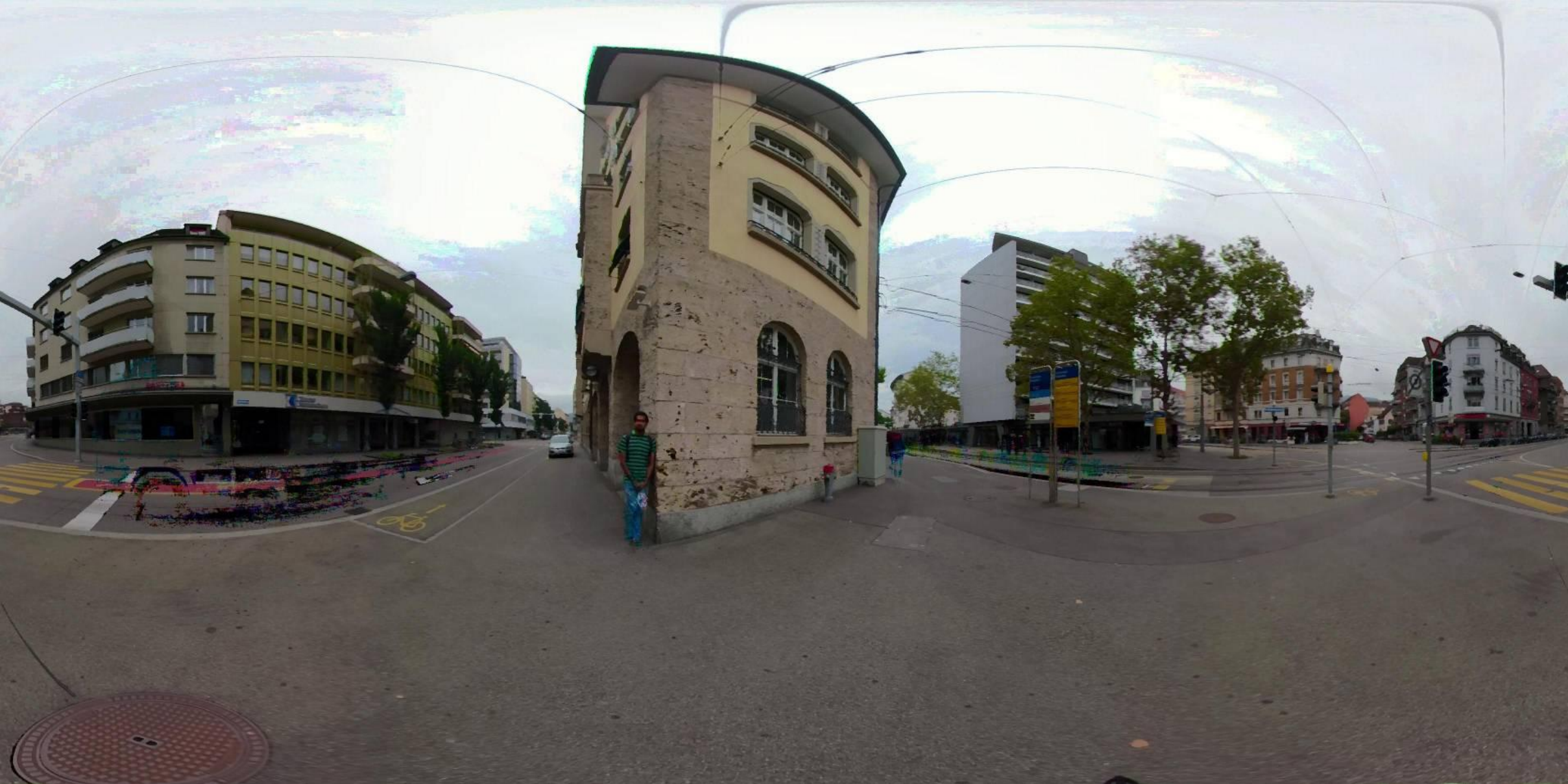}
    \hfil
    \includegraphics[width=0.24\textwidth]{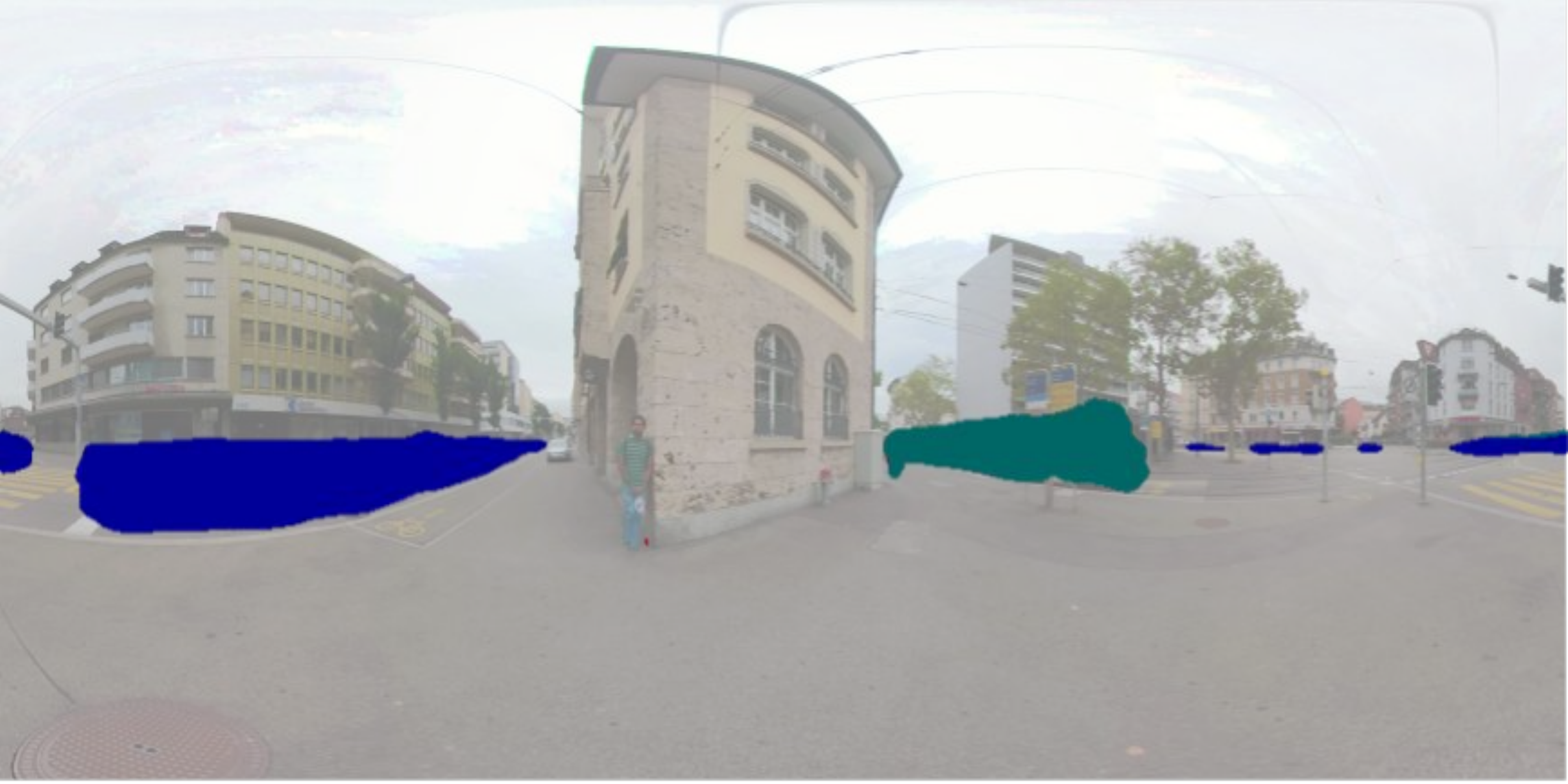}
    \hfil
    \includegraphics[width=0.24\textwidth]{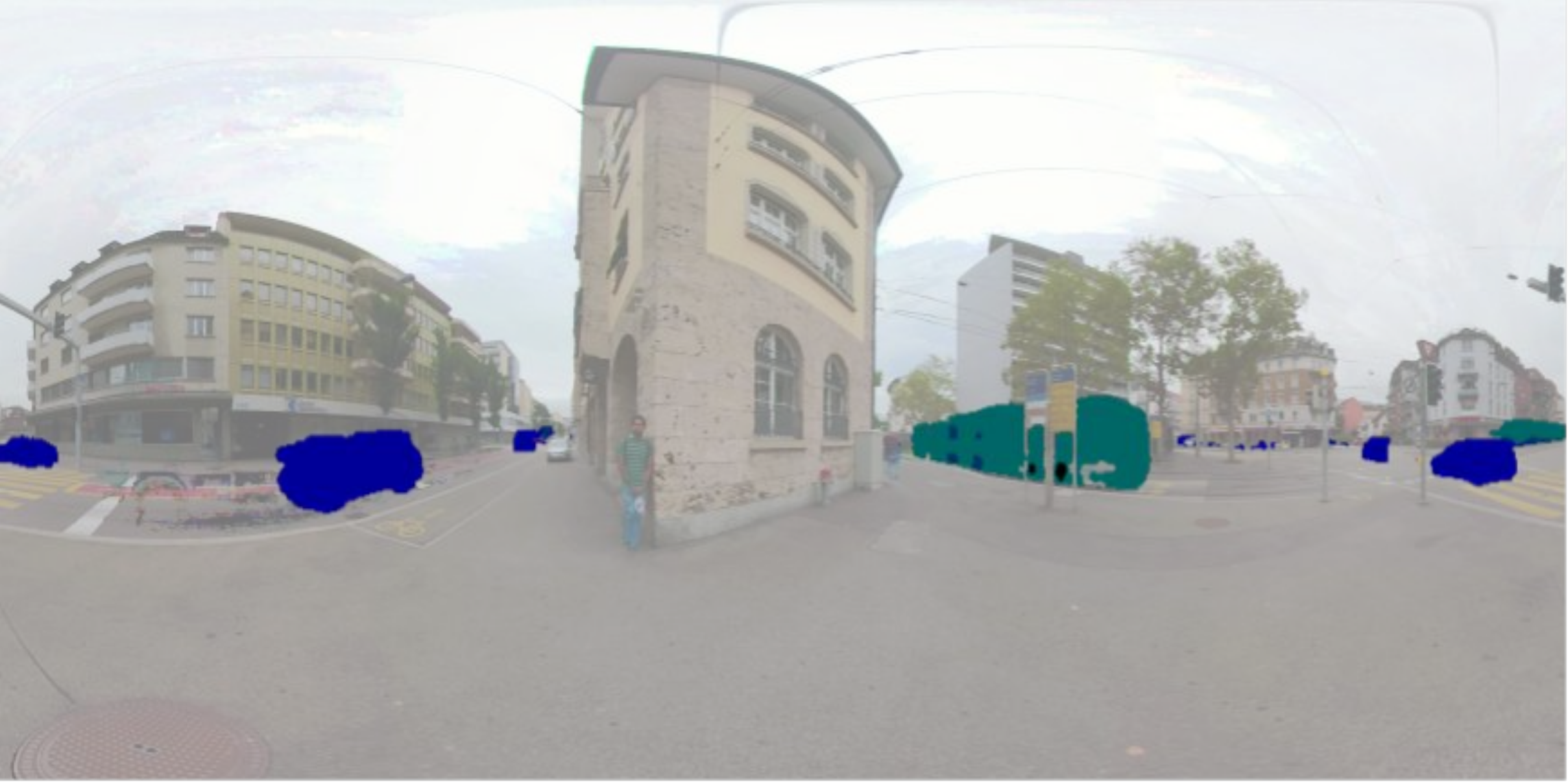}
    \\ \vspace{1mm}
    \includegraphics[trim=0 30 0 0,clip,width=0.24\textwidth,height=0.12\textwidth]{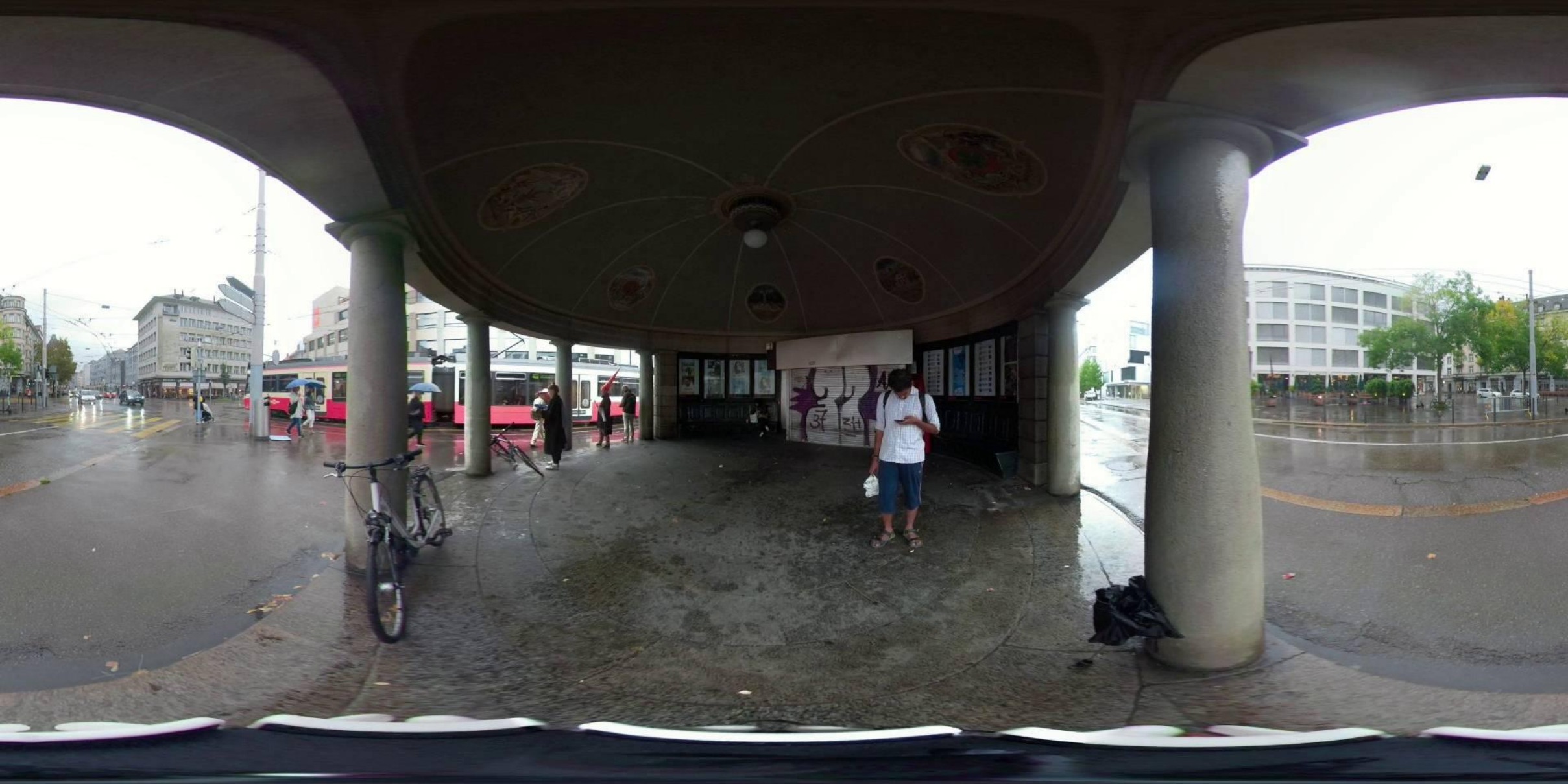}
    \hfil
    \includegraphics[trim=0 30 0 0,clip,width=0.24\textwidth,height=0.12\textwidth]{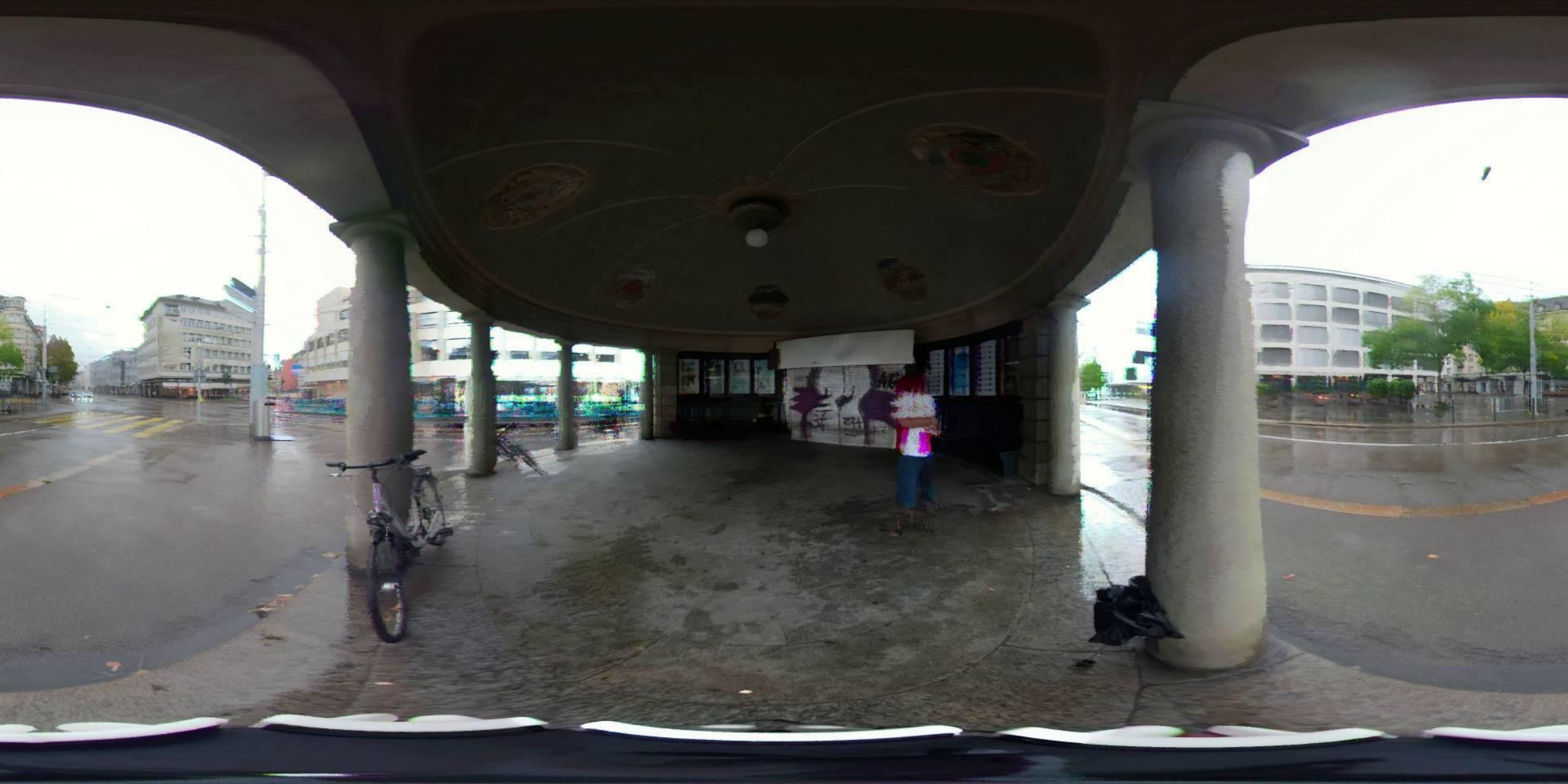}
    \hfil
    \includegraphics[width=0.24\textwidth]{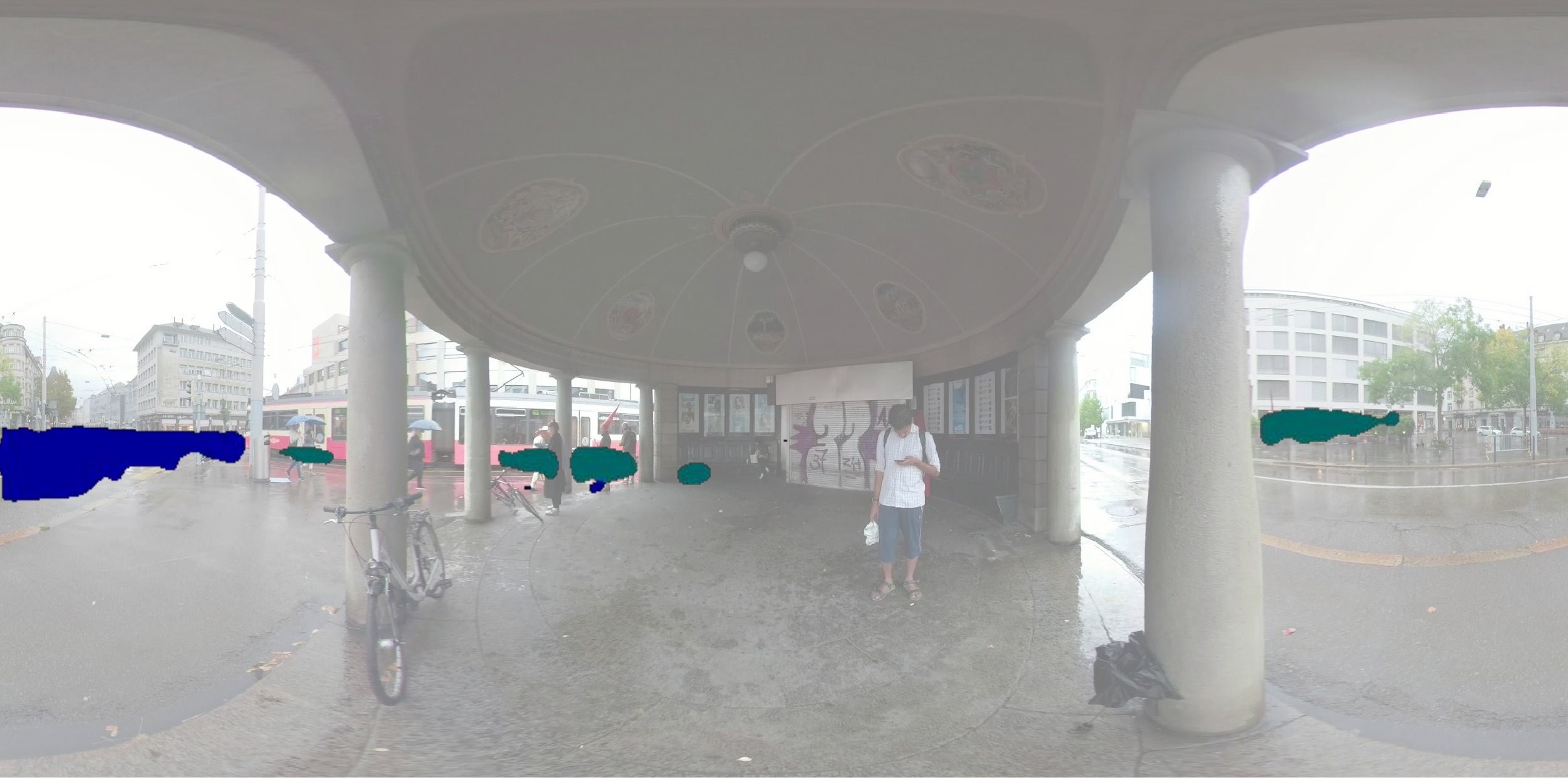}
    \hfil
    \includegraphics[width=0.24\textwidth]{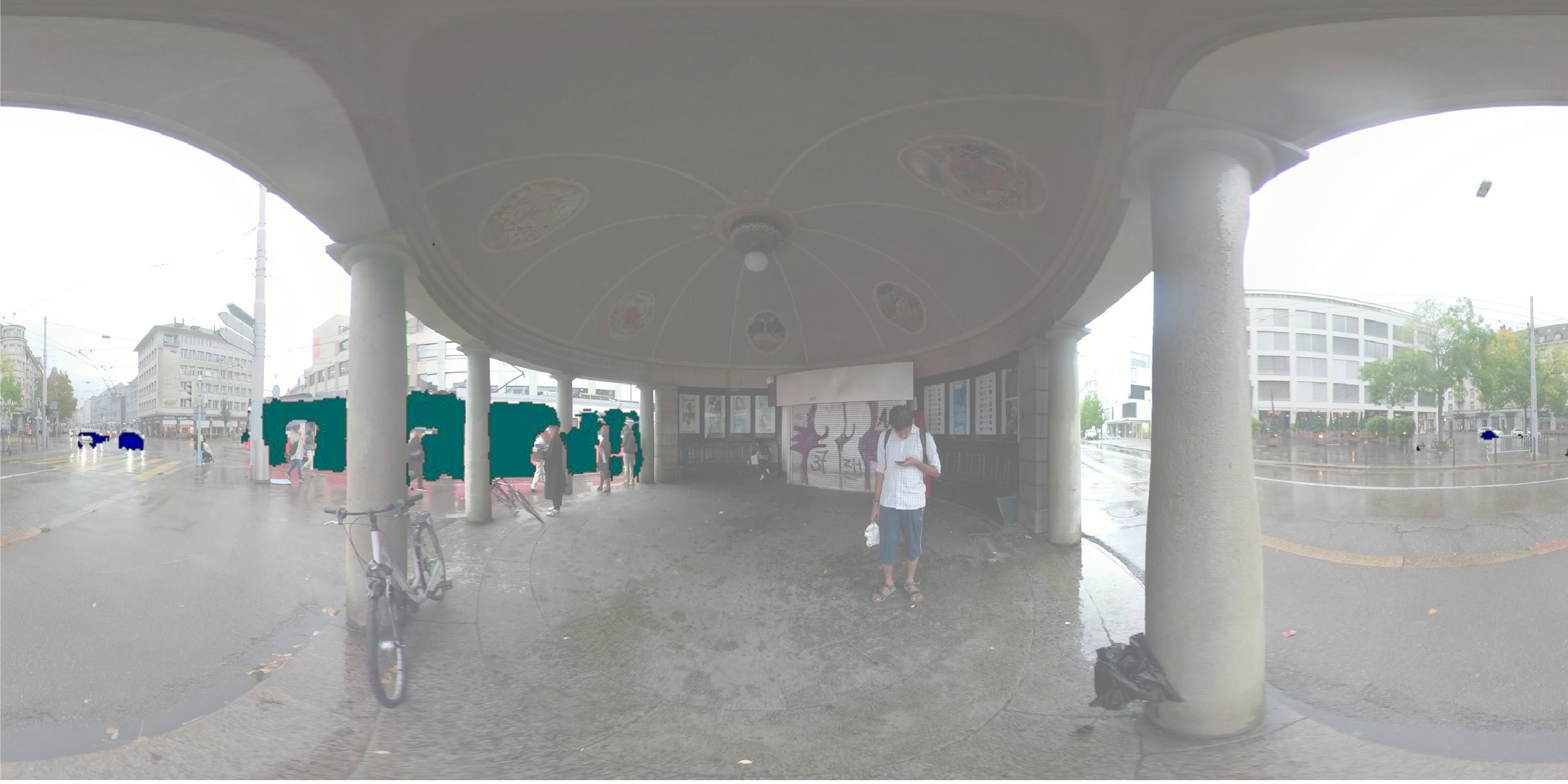}
    \\ \vspace{1mm}
    \includegraphics[trim=0 40 0 0,clip,width=0.24\textwidth,height=0.12\textwidth]{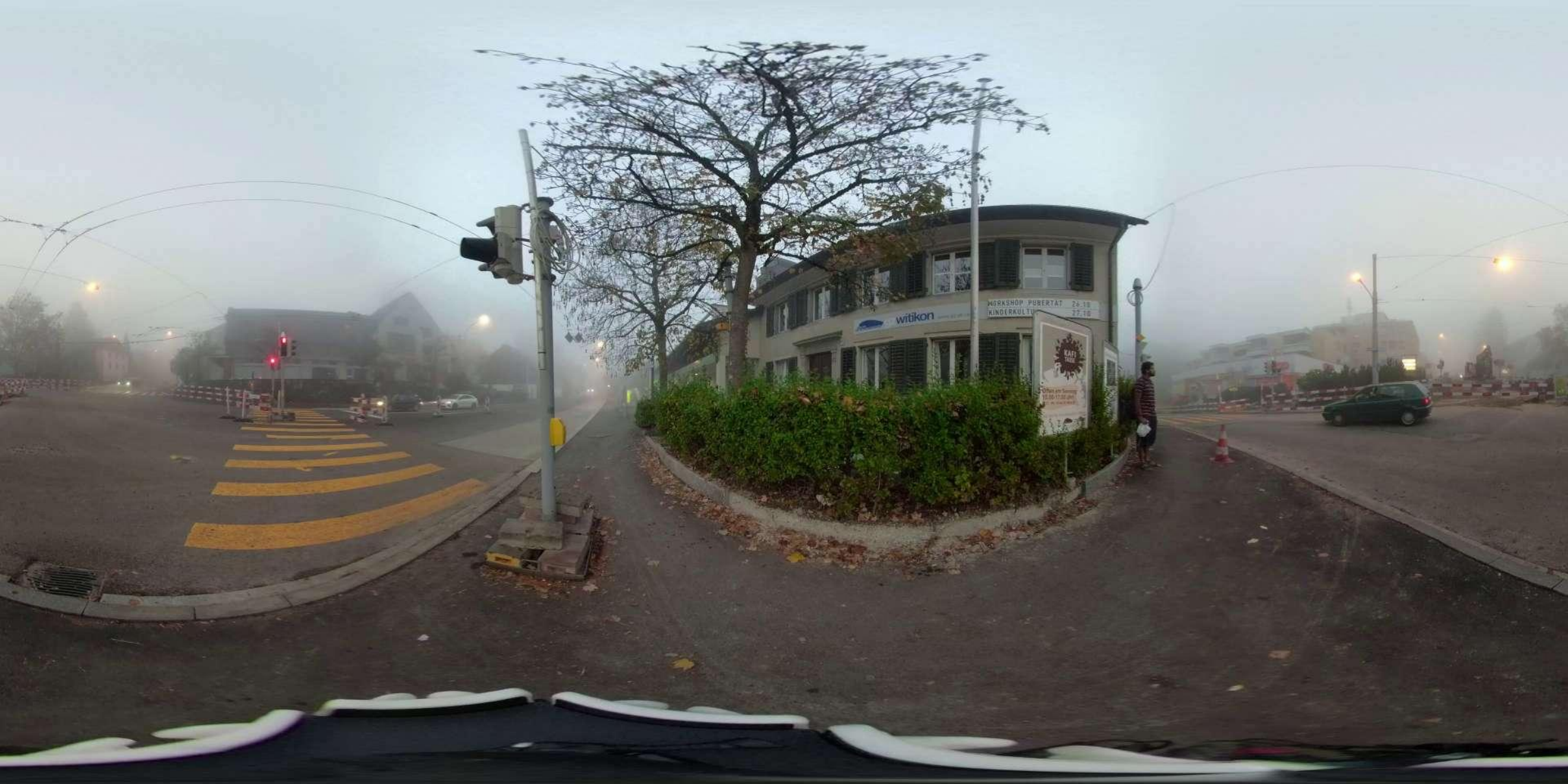}
    \hfil
    \includegraphics[trim=0 40 0 0,clip,width=0.24\textwidth,height=0.12\textwidth]{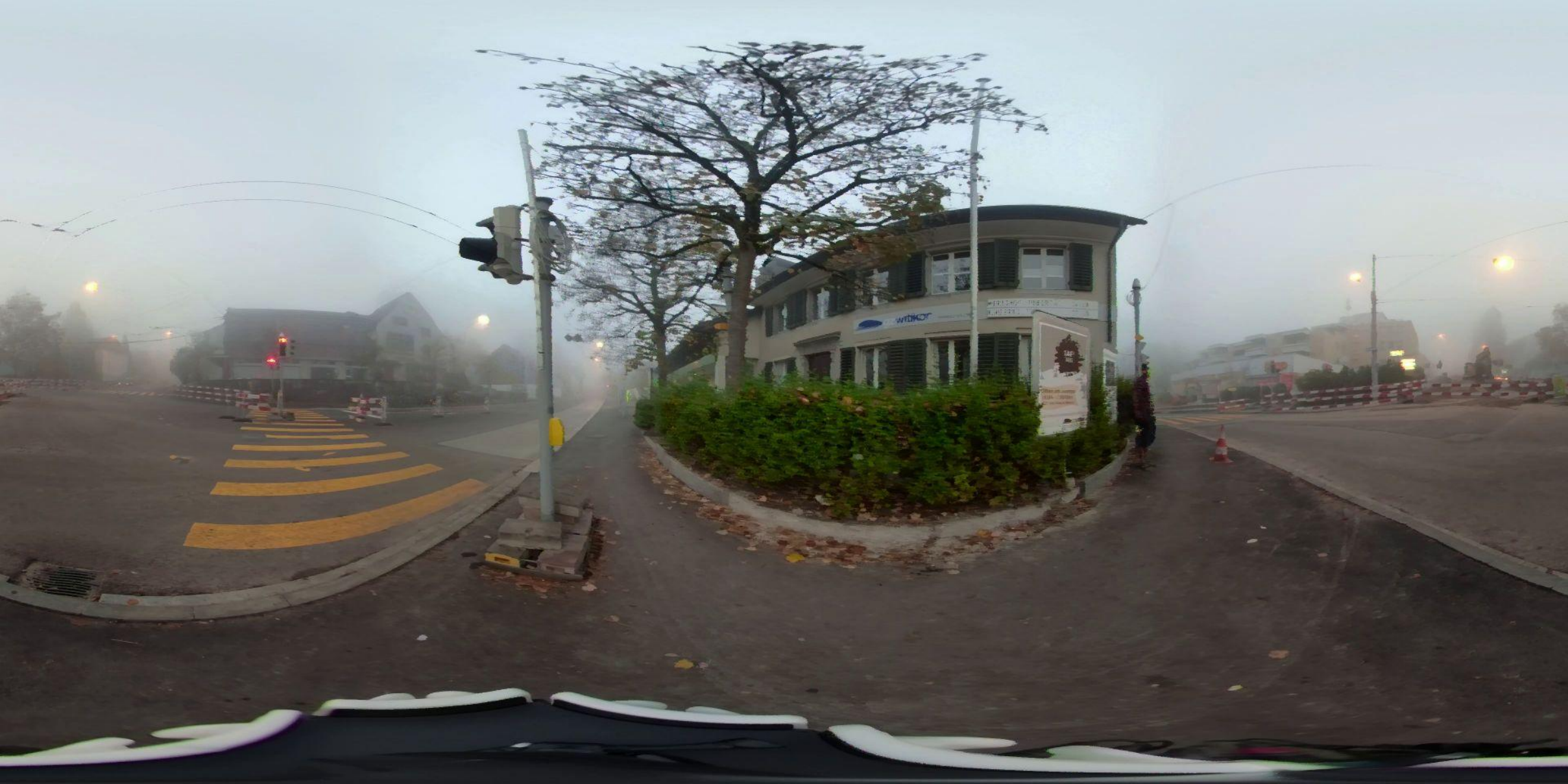}
    \hfil
    \includegraphics[width=0.24\textwidth]{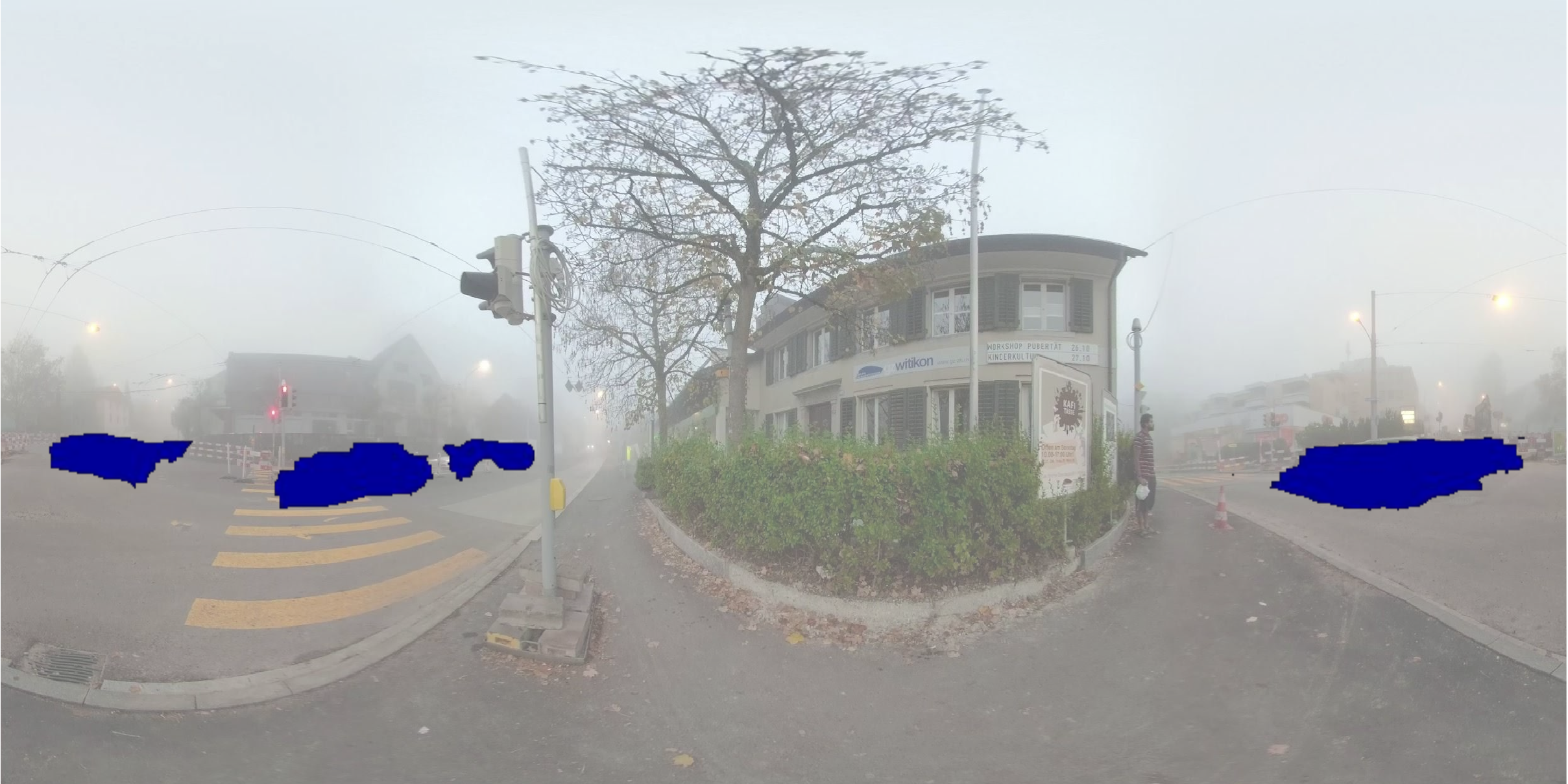}
    \hfil
    \includegraphics[width=0.24\textwidth]{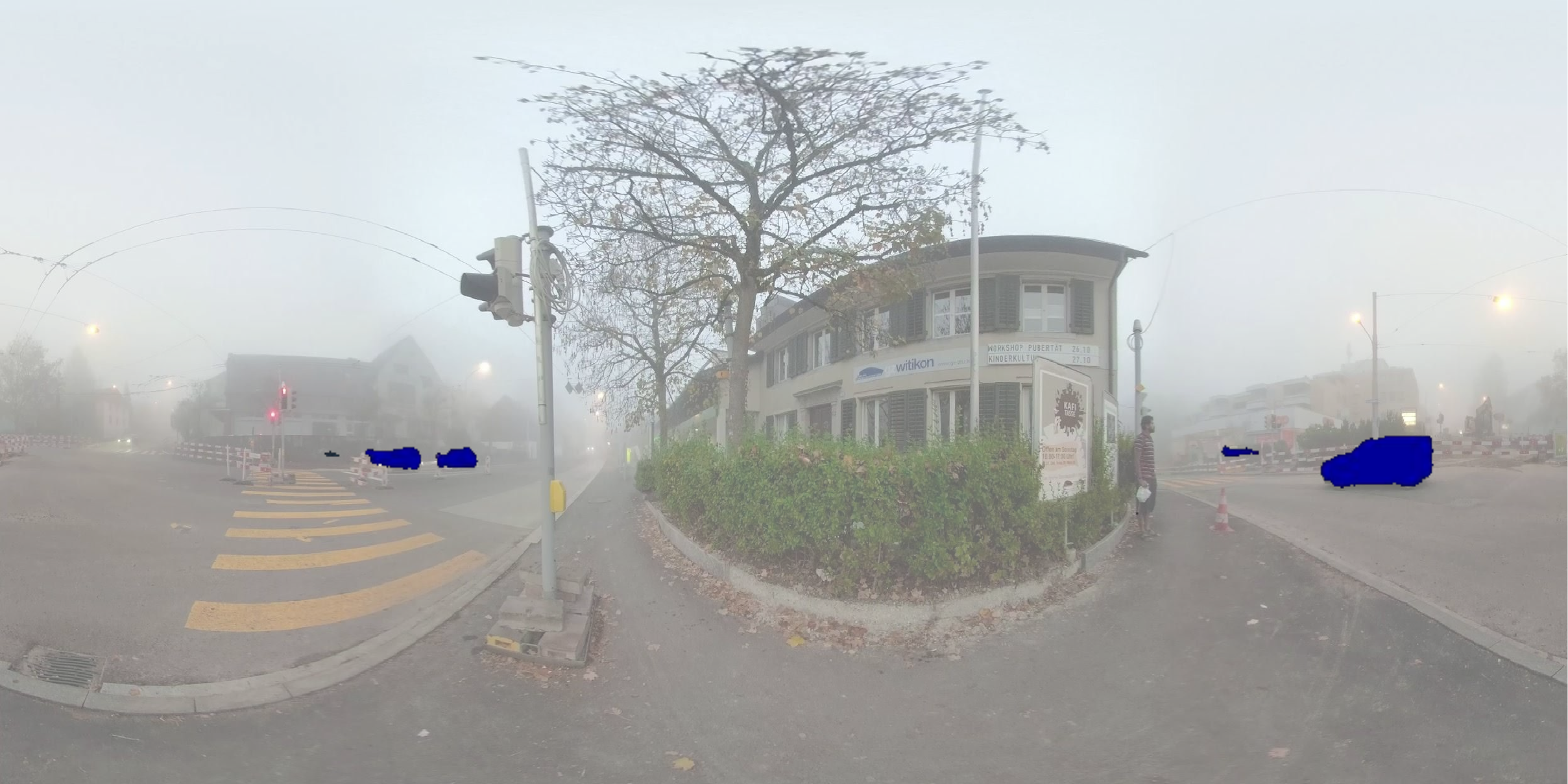}
    \\ \vspace{1mm}
    \includegraphics[trim=0 35 0 0,clip,width=0.24\textwidth,height=0.12\textwidth]{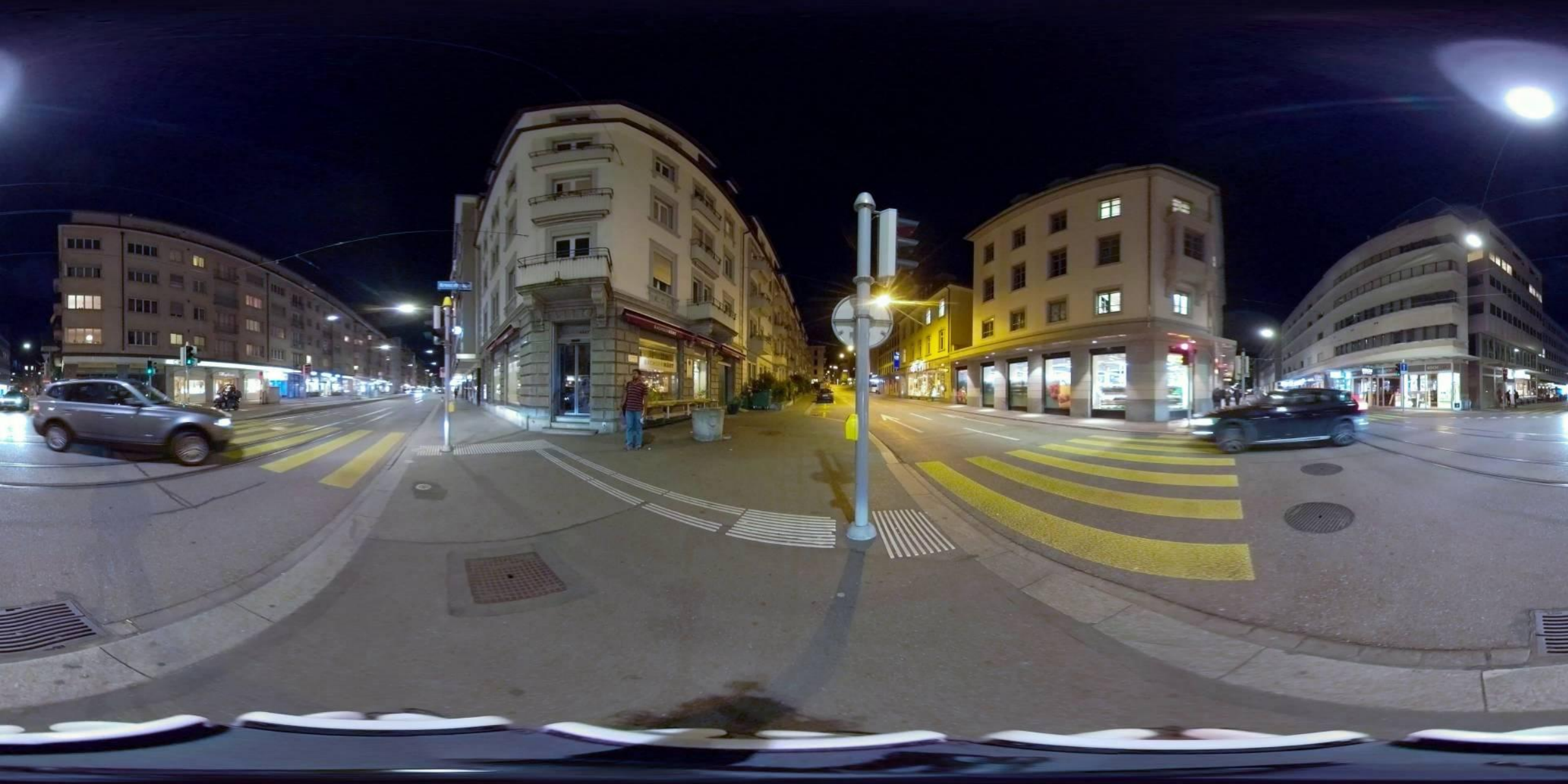}
    \hfil
    \includegraphics[trim=0 35 0 0,clip,width=0.24\textwidth,height=0.12\textwidth]{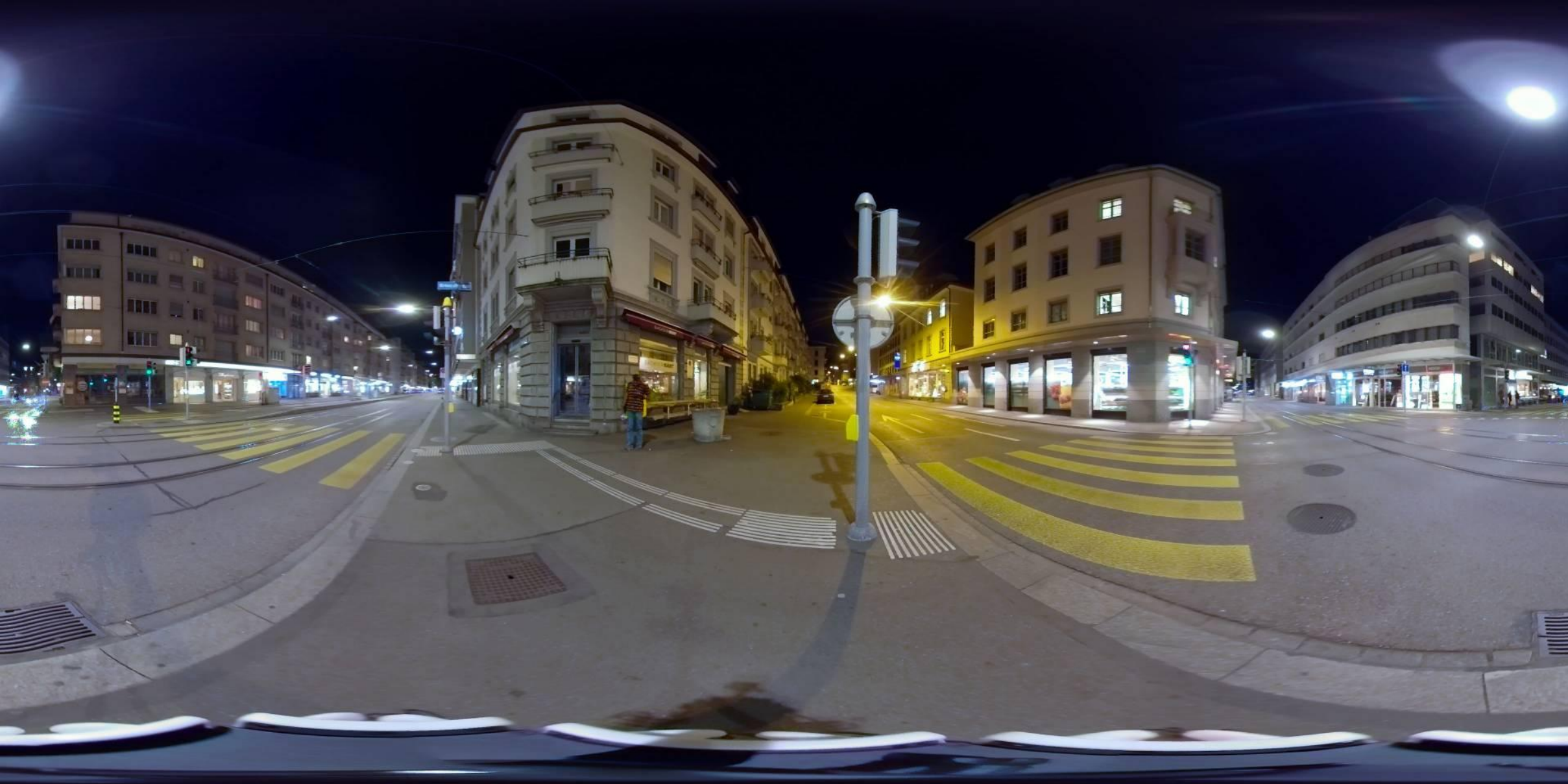}
    \hfil
    \includegraphics[width=0.24\textwidth]{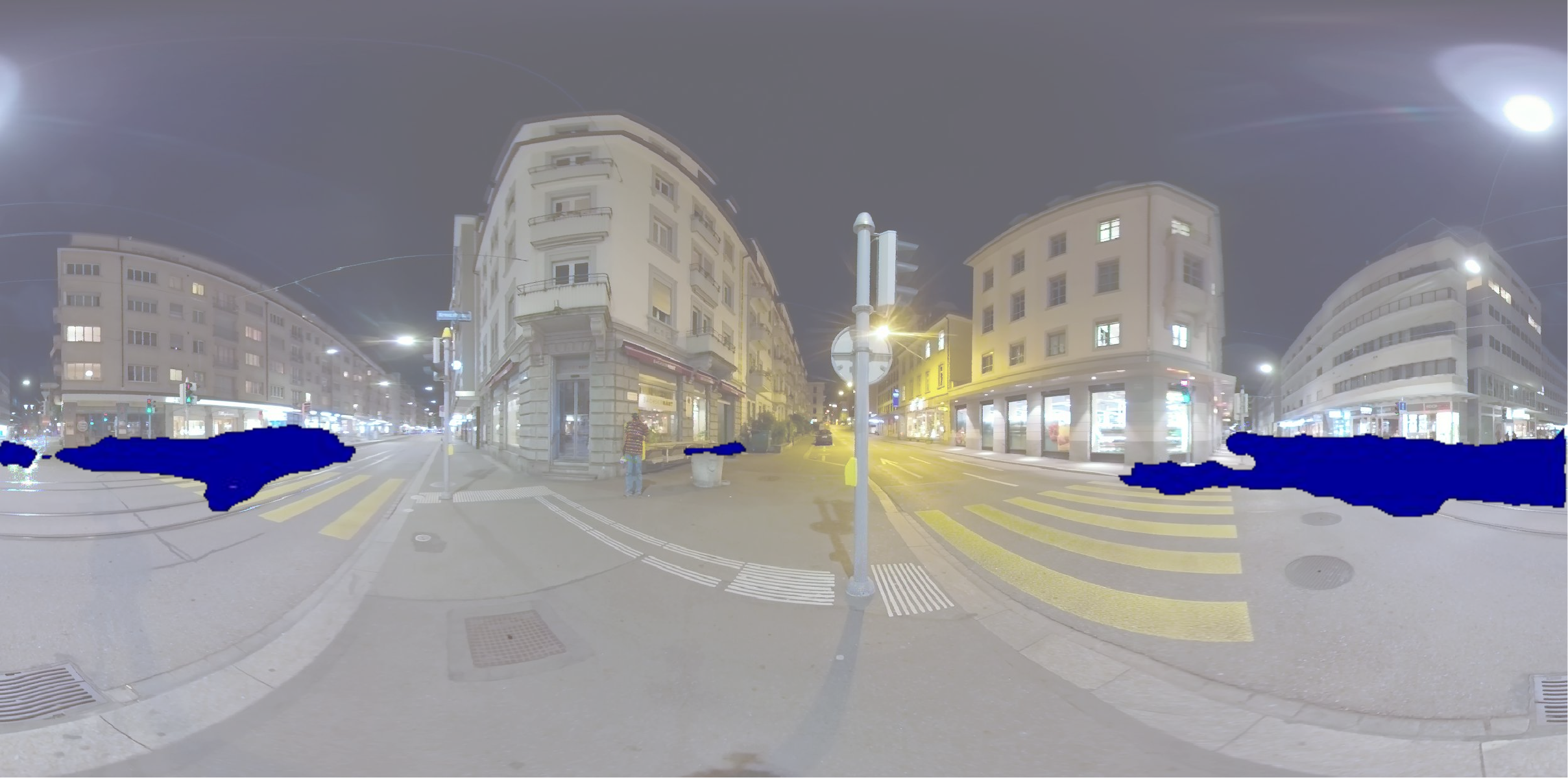}
    \hfil
    \includegraphics[width=0.24\textwidth]{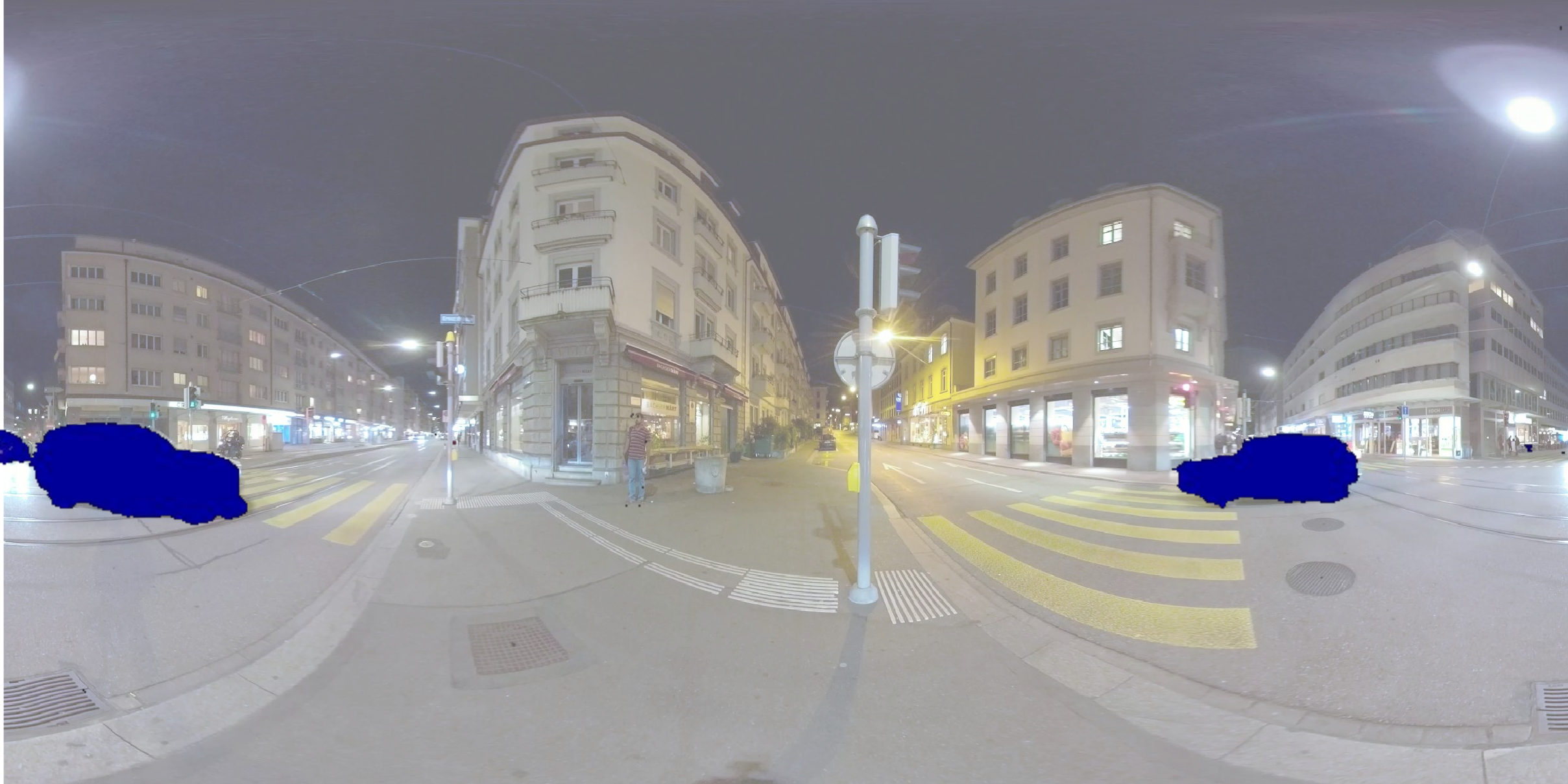}
    \\
    \hspace{-1.5mm} Visual Scene \hspace{4.8mm} Detected Background  \hspace{3mm}Semantic prediction \hspace{4.5mm}Semantic GT\hspace{-2.5mm}\\ 
    \caption{Qualitative results of auditory semantic prediction by our approach. The first column shows the visual scene, the second for the computed background image, the third for the semantic object masks predicted by our approach, and the fourth for the ground truth.}
    \label{fig:result1}\vspace{-2mm}
\end{figure*}


\smallskip
\noindent 
\textbf{Removing output channels degrades the performance}.
We vary the number of output microphone pairs for S$^3$R under the two multi-tasking models \emph{Ours(B:S)} and \emph{Ours(B:SD)}. We fix the input to pair (3,8) and experiment with different number of output pairs, ranging from 1 to 3. The results are presented in Fig.~\ref{fig:mic_ablation}(c) under these settings.
We see that spatial sound resolution to 3 binaural pairs performs better than to  1 or 2 binaural pairs. The more output channels we have, the better the semantic prediction results are.



\smallskip
\noindent
\textbf{ASPP is a powerful audio encoder}. 
We have found in our experiments that ASPP is a powerful encoder for audio as well. 
We compare our audio encoder with and without the ASPP module. For instance, Mono sound with ASPP clearly outperforms itself without ASPP -- adding ASPP improves the performance from 13.21 to 22.12  for mean IoU. The same trend is observed for other cases.


\begin{table}[!tb]
  \centering 
  \floatbox[\capbeside]{table}[0.7\textwidth]
{\caption{Depth prediction results with Mono and binaural sounds under different multi-task settings. For all the metrics, lower score is better.}\label{tab:depthprediction}}%
  {\begin{tabular}{ccccccccccc}
\toprule
  \multicolumn{2} {c} {Microphone} & \multicolumn{2}{c}{Joint Tasks} &  \multicolumn{4}{c}{Metrics} \\
 Mono & Binaural & Semantic & S$^3$R  & Abs Rel & Sq Rel & RMSE & MSE \\  \midrule
 \cmark& & & & 118.88 & 622.58 & 5.413& 0.365 \\
 &\cmark & & & 108.59 & 459.69 & 5.263& 0.331 \\
 &\cmark& & \cmark & 90.43 & 400.53 & 5.193 & 0.318 \\ 
 &\cmark&\cmark & & 87.96 & 290.06 & 5.136 & 0.315 \\
 &\cmark&\cmark & \cmark &  \textbf{84.24} &  \textbf{222.41} &  \textbf{5.117} &  \textbf{0.310} \\
\bottomrule
\end{tabular}} 

\end{table} 

\vspace{-2mm}
\subsection{Auditory depth prediction}
We report the depth prediction results in Tab.~\ref{tab:depthprediction}. 
The first row represents the depth prediction from Mono sounds alone while the second row represents our baseline method where the depth map is predicted from binaural sounds. As a simple baseline, we also compute the mean depth over the training dataset and evaluate over the test set. The RMSE and MSE scores are $15.024$ and $0.864$, respectively, which are 2.5 times worse than our binaural sound baseline. The multi-task learning with S$^{3}$R and semantic prediction under shared audio encoder also provides pronounced improvements for depth prediction. Jointly training all the three tasks yields the best performance for depth prediction as well. 

\begin{table}[!bt]
  \centering 
  \small
  \setlength\tabcolsep{3pt}
  \begin{adjustbox}{max width=\textwidth,max totalheight=\textheight}
  \begin{tabular}{ccccccccccc}
\cmidrule[\heavyrulewidth]{1-8}
 \multicolumn{2}{c}{Joint Tasks} &  \multicolumn{2}{c}{Mic Ids} &  \multicolumn{4}{c}{Metrics} & &\multirow{6}{*}{
 \includegraphics[trim=35 0 0 55,clip,width=0.4\textwidth,height=0.24\textwidth]{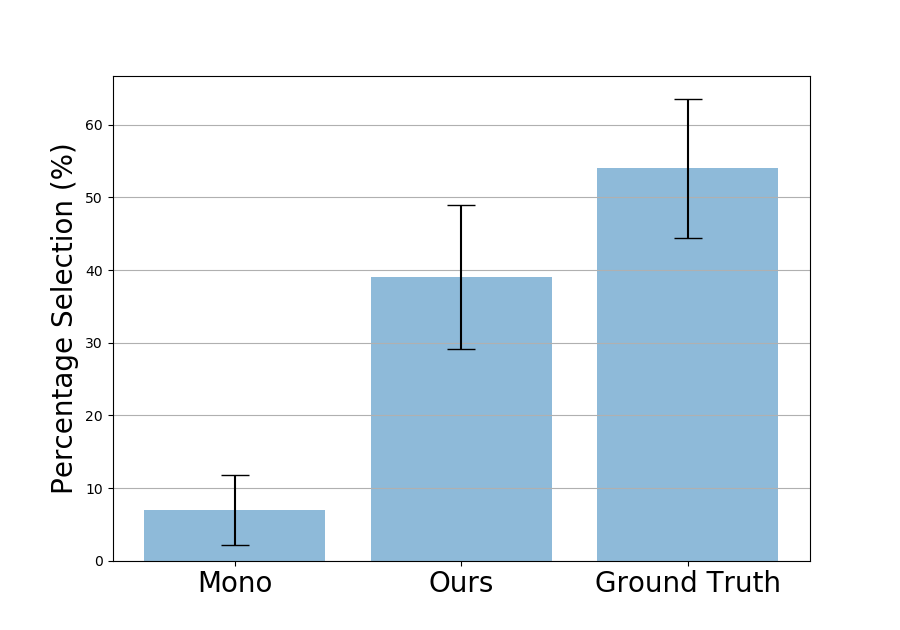} \vspace{-3.5mm} }\\ 
 Semantic & Depth & In & Out & MSE-1   & ENV-1  & MSE-2   & ENV-2 &\\  
 \cmidrule(lr){1-8}
 & & (3,8) & (1,6) &0.1228 & 0.0298 &0.1591 & 0.0324&\\ 
  & \cmark & (3,8) & (1,6) & 0.0984 & 0.0221 & 0.1044 & 0.0267 &\\
    \cmark  & \cmark & (3,8) & (1,6) & 0.0978 & 0.0218 & 0.1040 & 0.0264 &\\
 \cmark & & (3,8) & (1,6) &  \textbf{0.0956} & \textbf{0.0214} & \textbf{0.1001} & \textbf{0.0243} &\\
\cmidrule[\heavyrulewidth]{1-8} \vspace{-2mm}\\

& & \multicolumn{4}{c} (a) S$^{3}$R results in multi-tasking & & & & (b) Subjective evaluation &\\
\end{tabular} \end{adjustbox} 
\caption{a) S$^3$R results. MSE1 and MSE2 represent mean squared error while ENV1 and ENV represent envelope error for the 2 output channels of binaural sounds. For all metrics, lower score is better. b) Subjective assessment of the generated binaural sounds.} \vspace{-2mm}
  \label{tab:spatialresolutionresults}
\end{table} 


\vspace{-2mm}
\subsection{Spatial Sound Super-resolution} 
Tab.~\ref{tab:spatialresolutionresults} shows the results of S$^3$R as a stand-alone task (first row) and under the multi-task setting. To keep it simple, we estimate the sound signals of microphone pair $(1,6)$ alone from the microphone pair $(3,8)$. We can see from Fig.~\ref{fig:setup} that these two pairs are perpendicular in orientation, so the prediction is quite challenging. 
We can observe from Tab.~\ref{tab:spatialresolutionresults} that the multi-task learning with semantic prediction task and depth prediction task outperforms the accuracy of the stand-alone S$^{3}$R model. Hence, the multi-task learning also helps S$^{3}$R -- the same trend as for semantic perception and depth perception.  We also conduct a user study for the subjective assessment of the generated binaural sounds.
The participants listen to ground-truth binaural sounds, binaural sounds generated from \emph{Mono} approach (Tab.~\ref{tab:maintable}) and from our approach. We present two (out of three) randomly picked sounds and ask the user to select a preferred one in temrs of binaural sound quality. Tab.~\ref{tab:spatialresolutionresults}(b) shows the percentage of times each method is chosen as the preferred one. We see that \textit{Ours} is close to the ground truth selection implying that our predicted binaural sounds are of high quality.

\begin{figure*}[tb]
\begin{tabular}{cccc}
  \centering 
  \includegraphics[trim=0 0 0 0,clip,width=0.235\textwidth,height=0.12\textwidth]{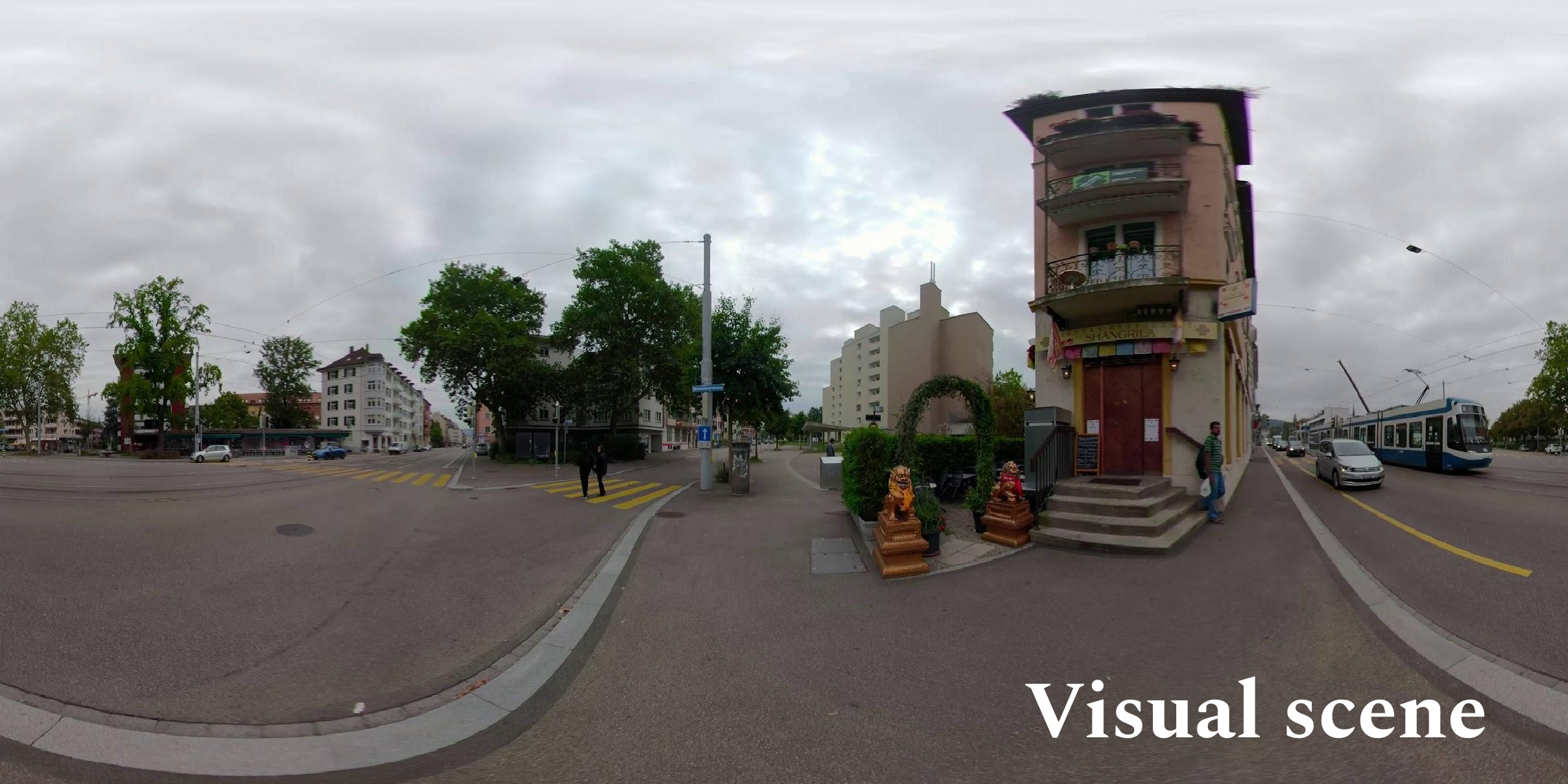} & 
  \includegraphics[trim=0 0 0 0,clip,width=0.235\textwidth,height=0.12\textwidth]{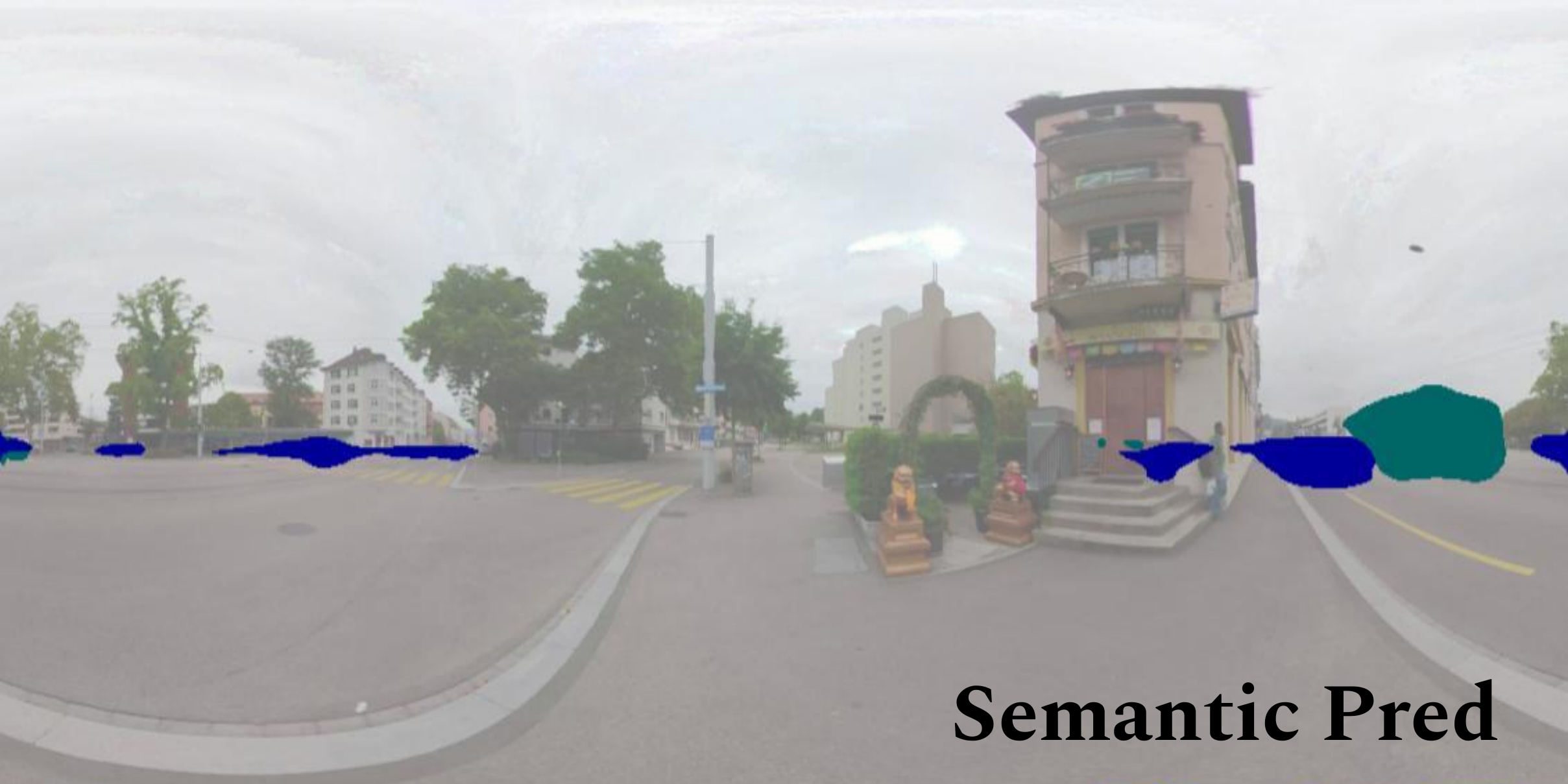} & 
  \includegraphics[trim=0 0 0 0,clip,width=0.235\textwidth,height=0.12\textwidth]{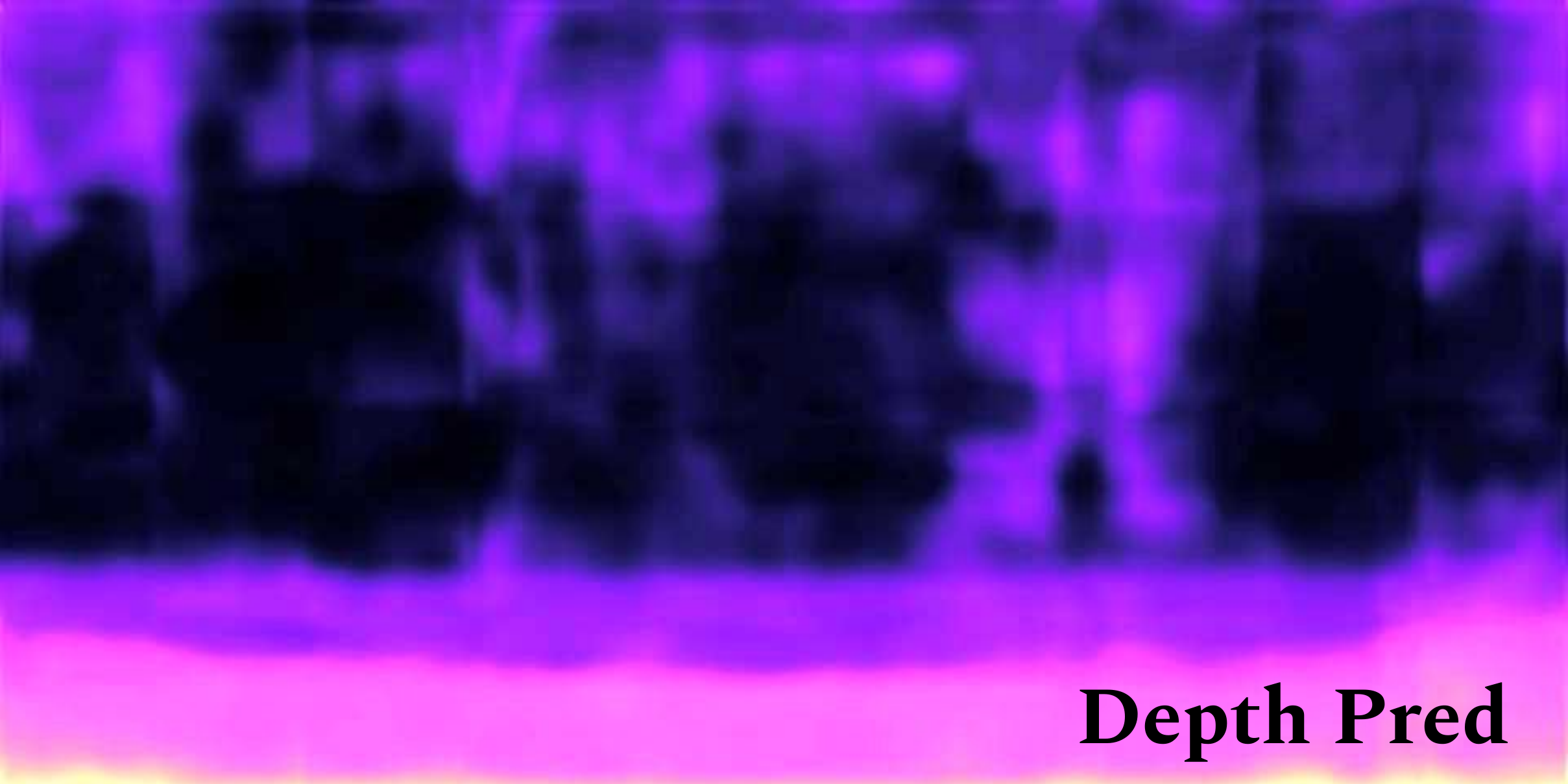} &
  \multirow{2}{*}[3.7em]{\includegraphics[trim=28 20 420 0,clip,width=0.275\textwidth,height=0.25\textwidth]{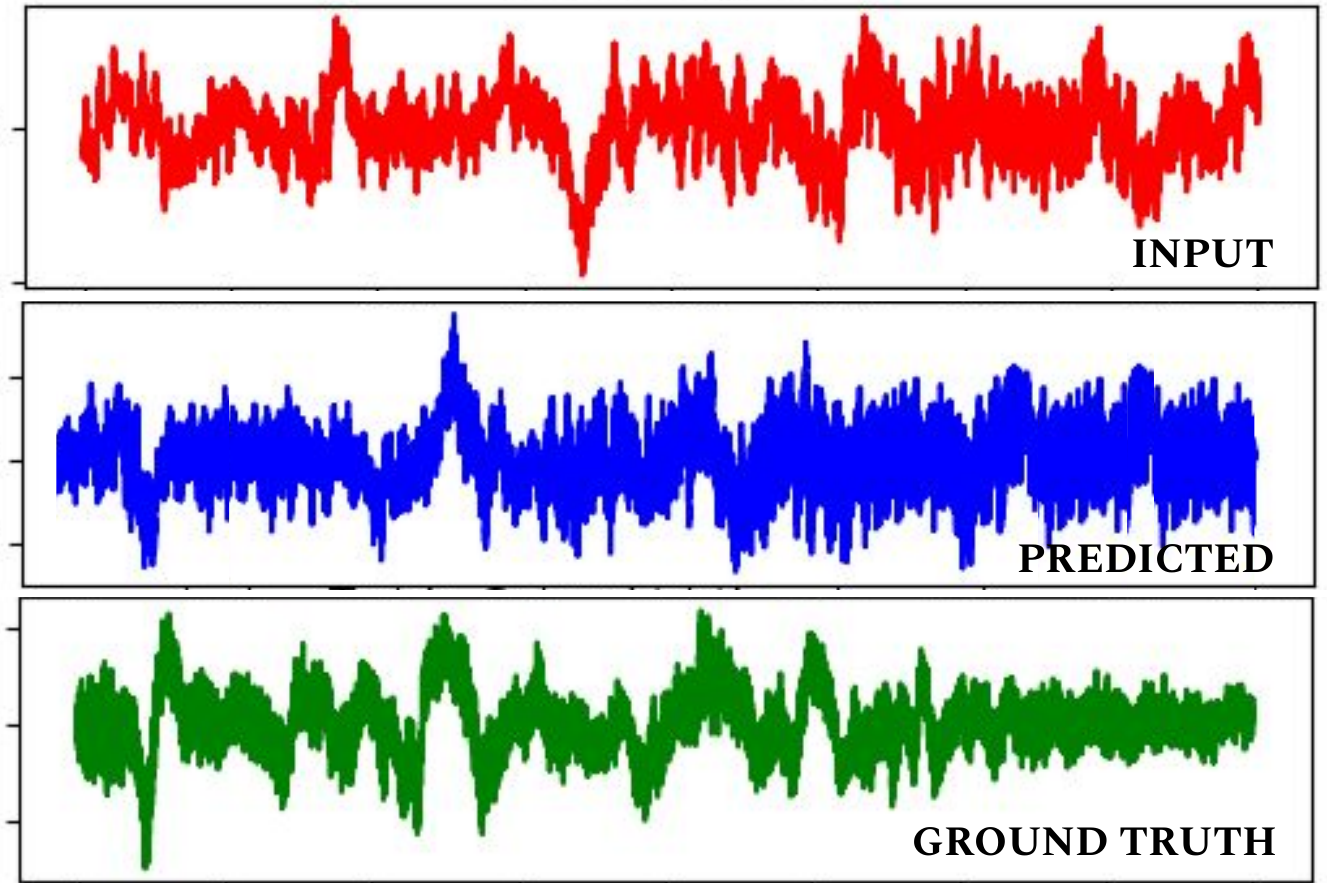}}
  \\
  \includegraphics[trim=0 0 0 0,clip,width=0.235\textwidth,height=0.12\textwidth]{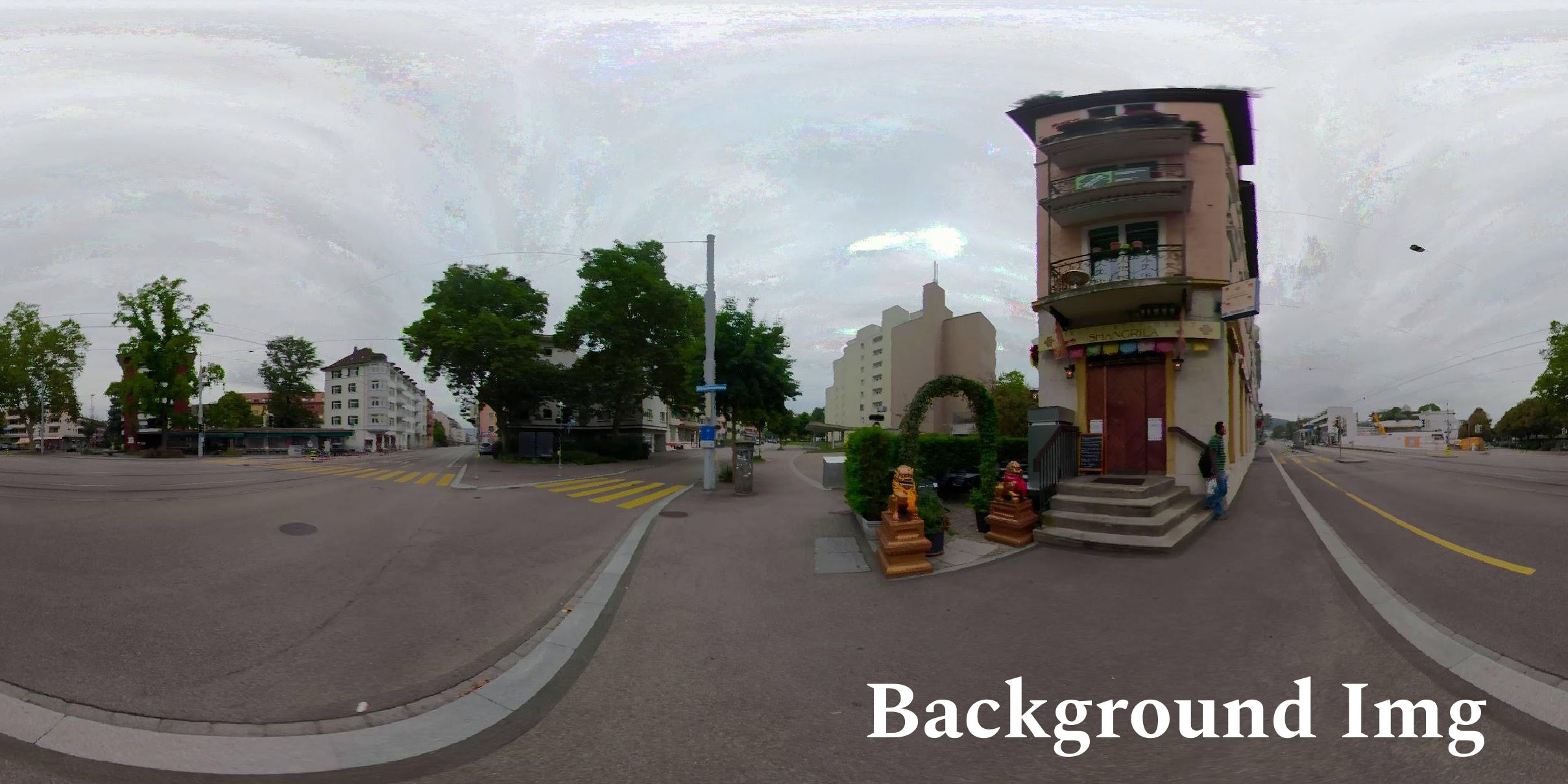} & 
  \includegraphics[trim=0 0 0 0,clip,width=0.235\textwidth,height=0.12\textwidth]{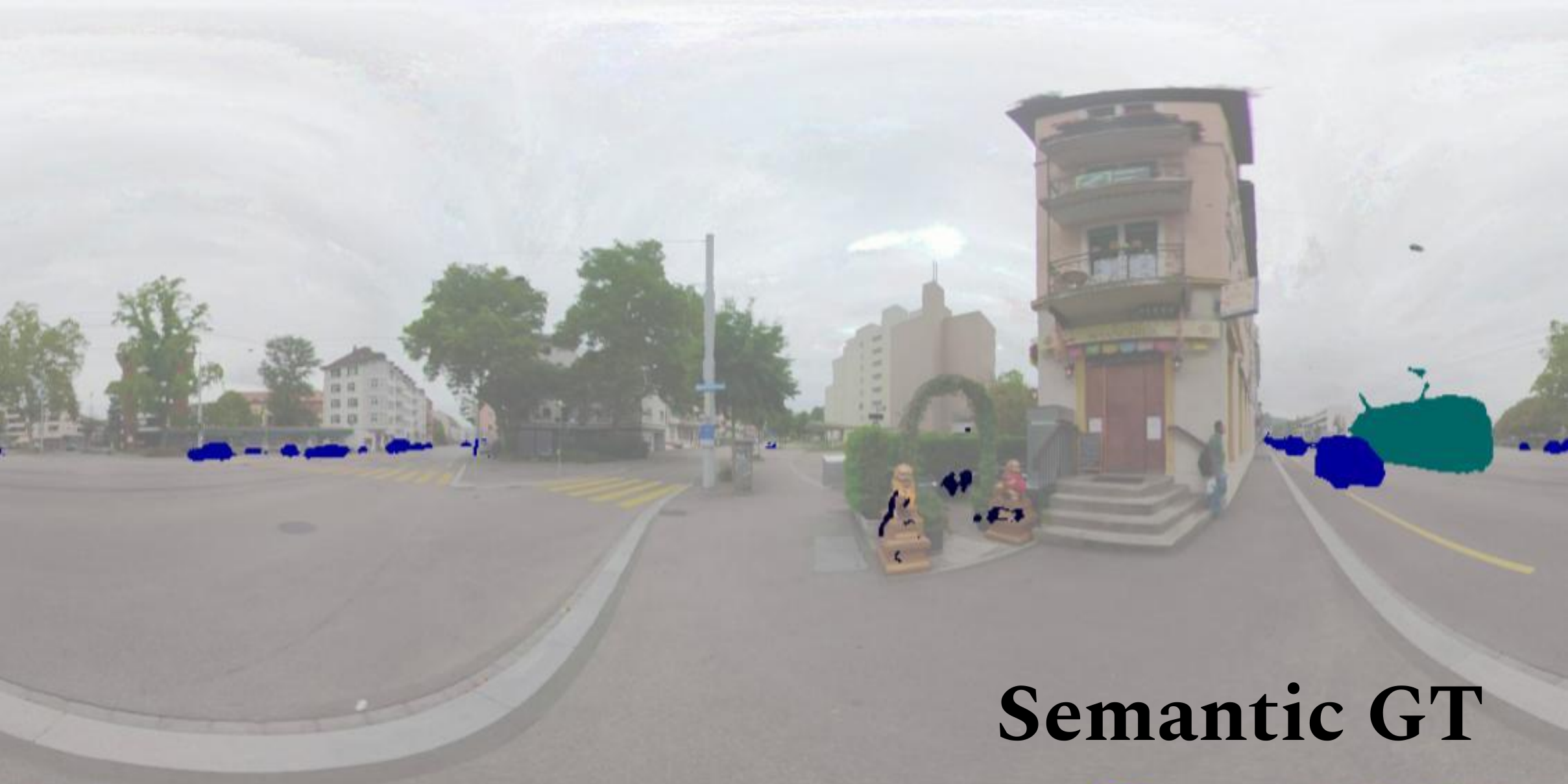} &
  \includegraphics[trim=0 0 0 0,clip,width=0.235\textwidth,height=0.12\textwidth]{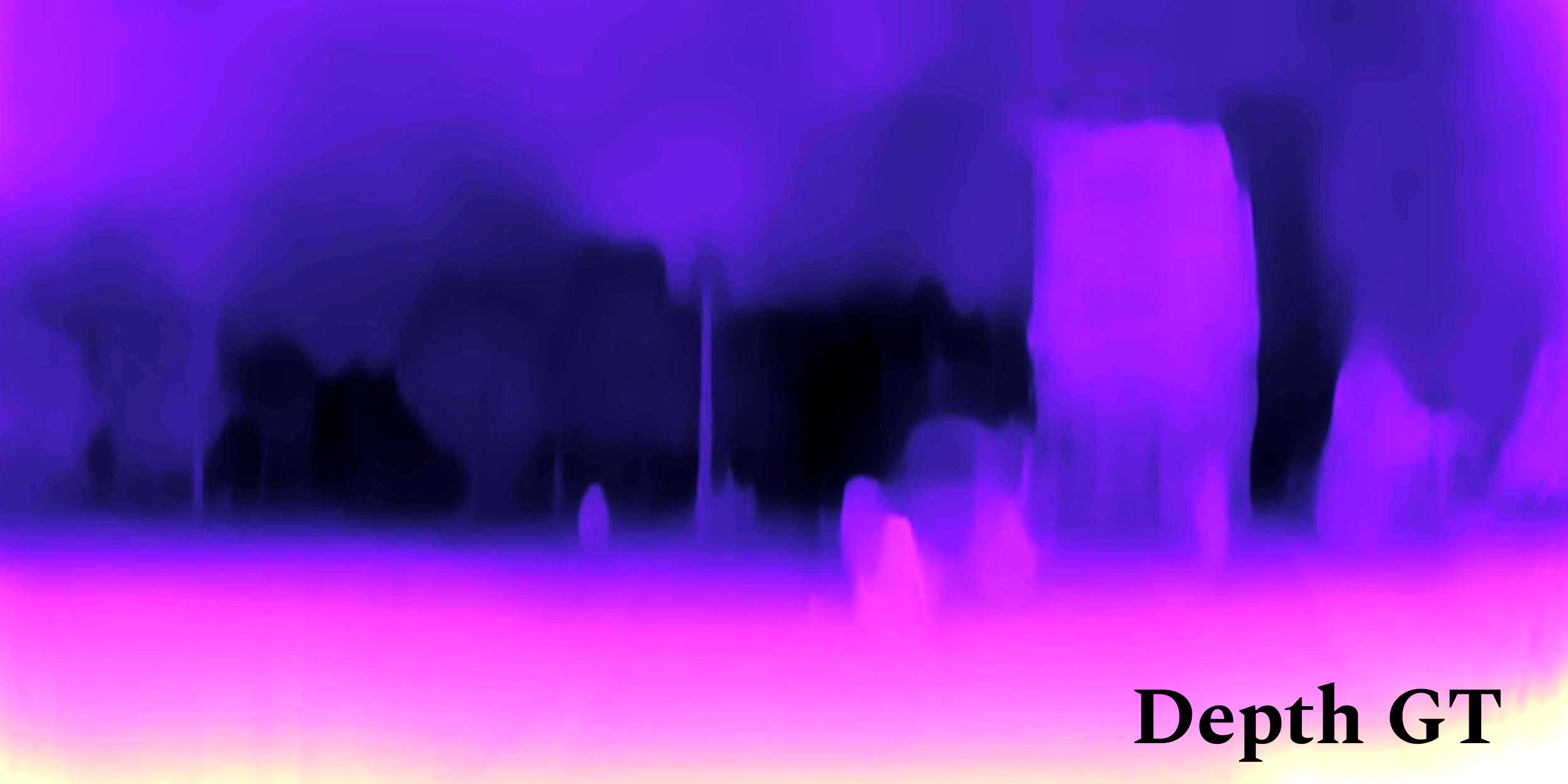} & 
  \\
  \includegraphics[trim=0 00 0 0,clip,width=0.235\textwidth,height=0.12\textwidth]{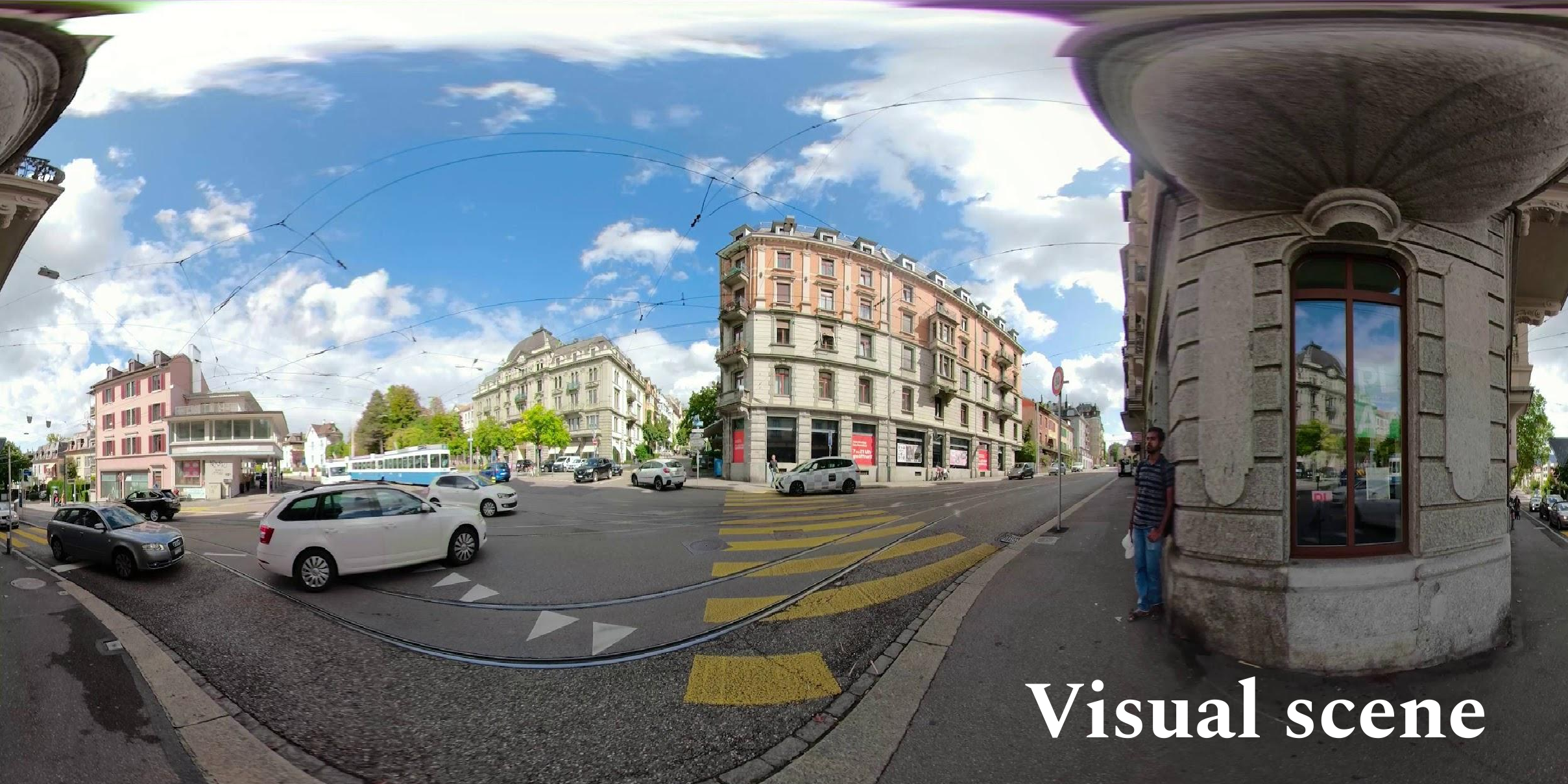} & 
  \includegraphics[trim=0 0 0 0,clip,width=0.235\textwidth,height=0.12\textwidth]{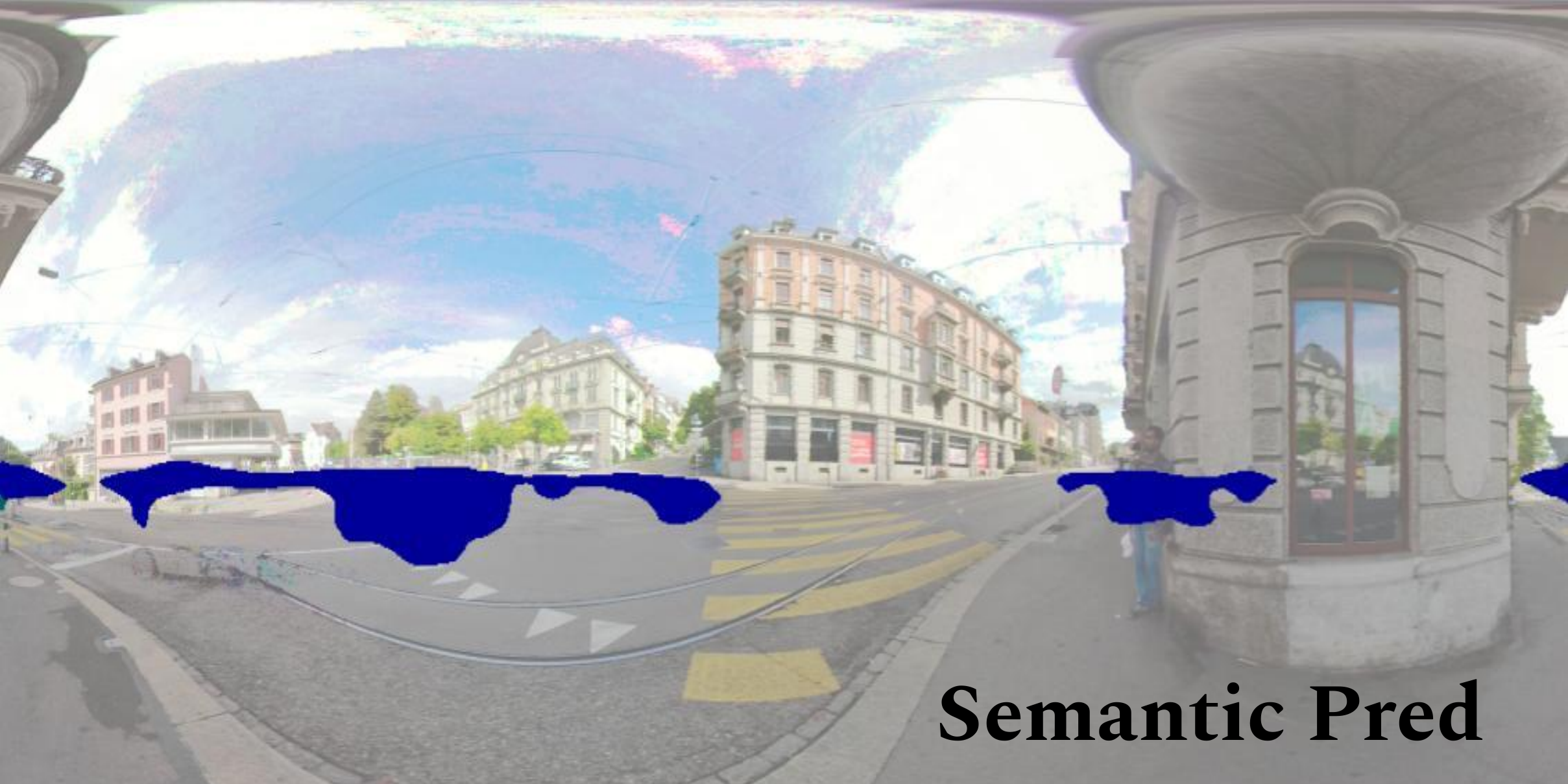} & 
  \includegraphics[trim=0 0 0 0,clip,width=0.235\textwidth,height=0.12\textwidth]{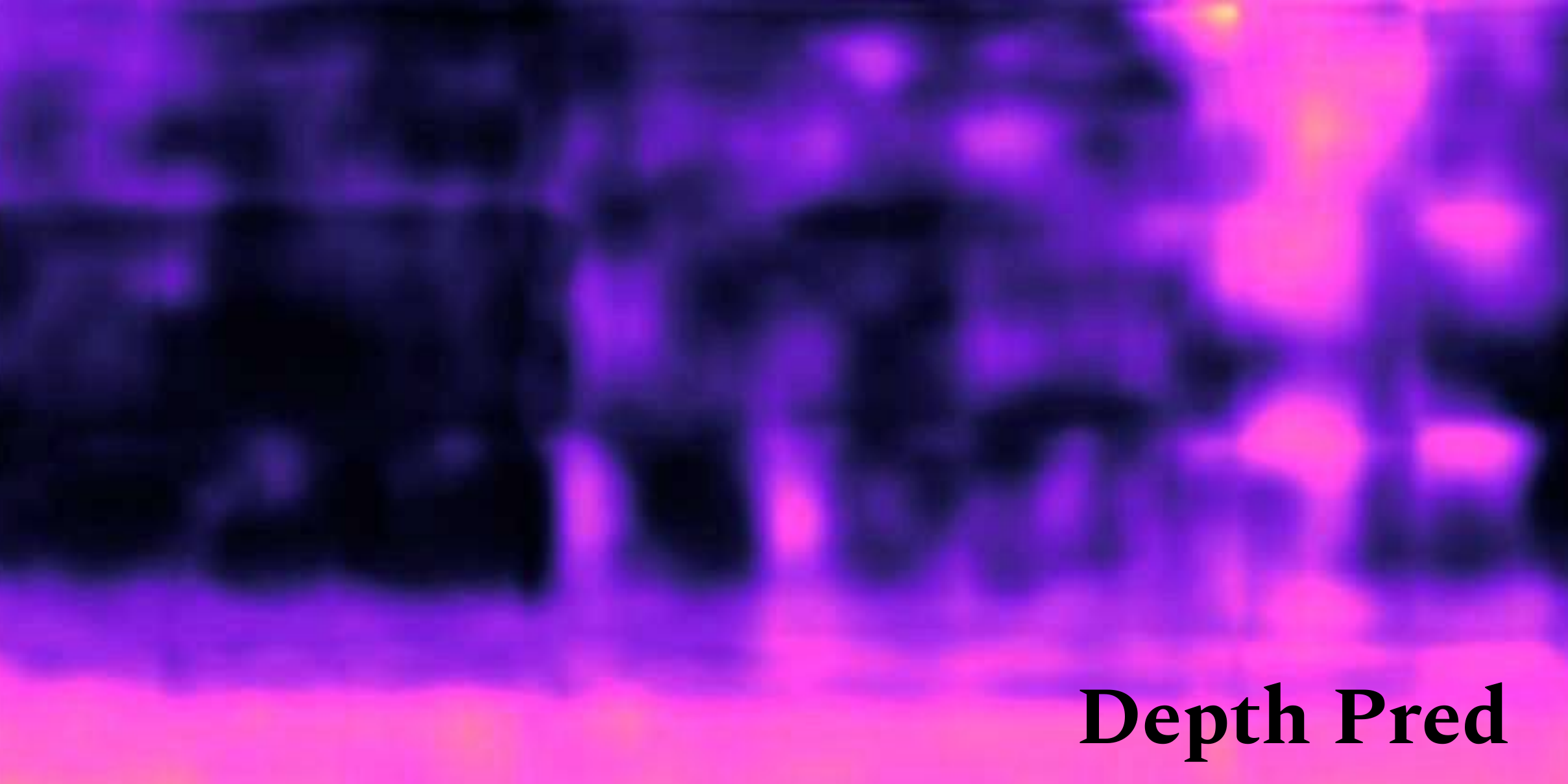} &
  \multirow{2}{*}[3.7em]{\includegraphics[trim=445 20 5 0,clip,width=0.275\textwidth,height=0.25\textwidth]{./images/sound_result.pdf}}
  \\
  \includegraphics[trim=0 0 0 0,clip,width=0.235\textwidth,height=0.12\textwidth]{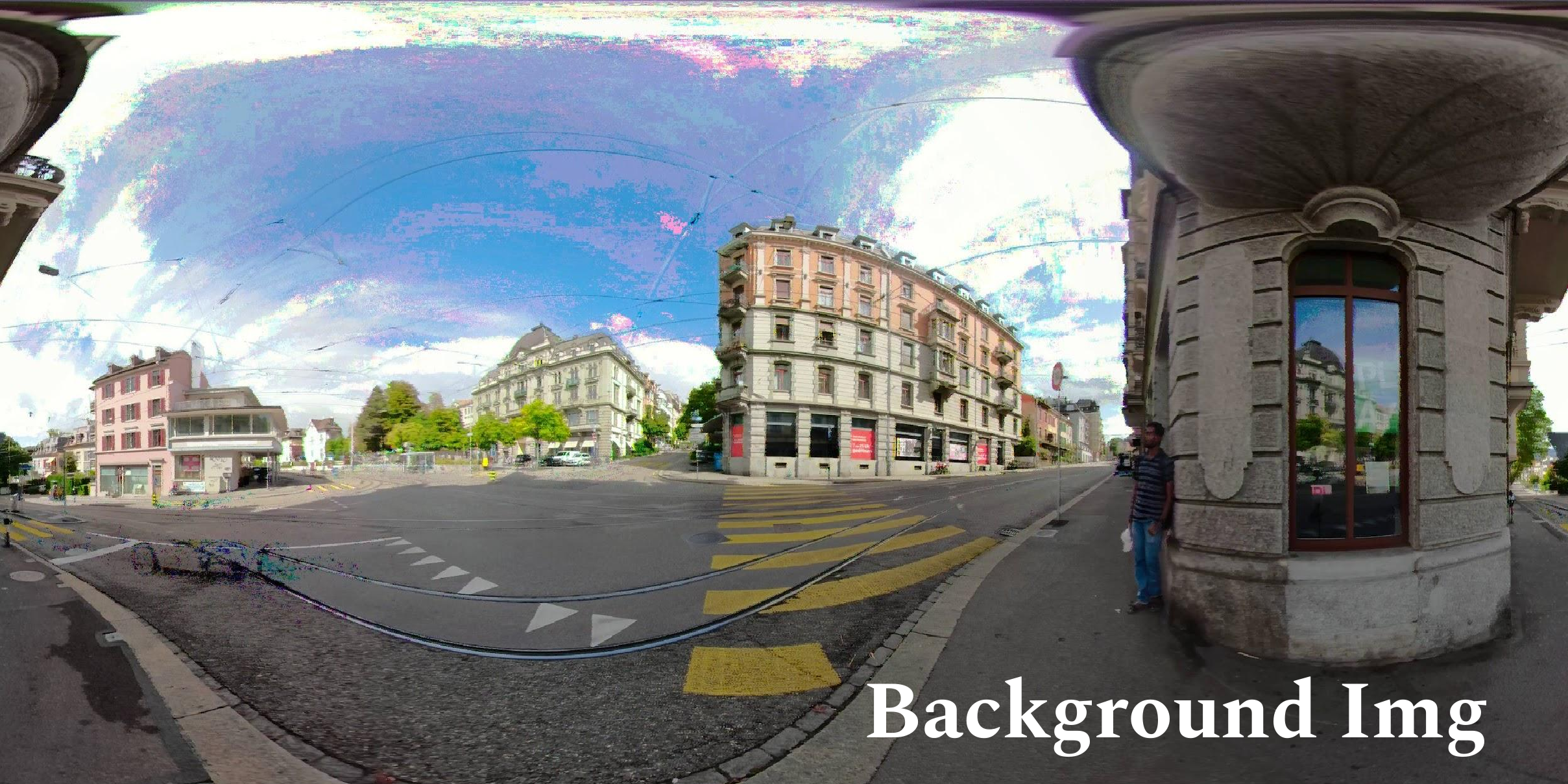} & 
  \includegraphics[trim=0 0 0 0,clip,width=0.235\textwidth,height=0.12\textwidth]{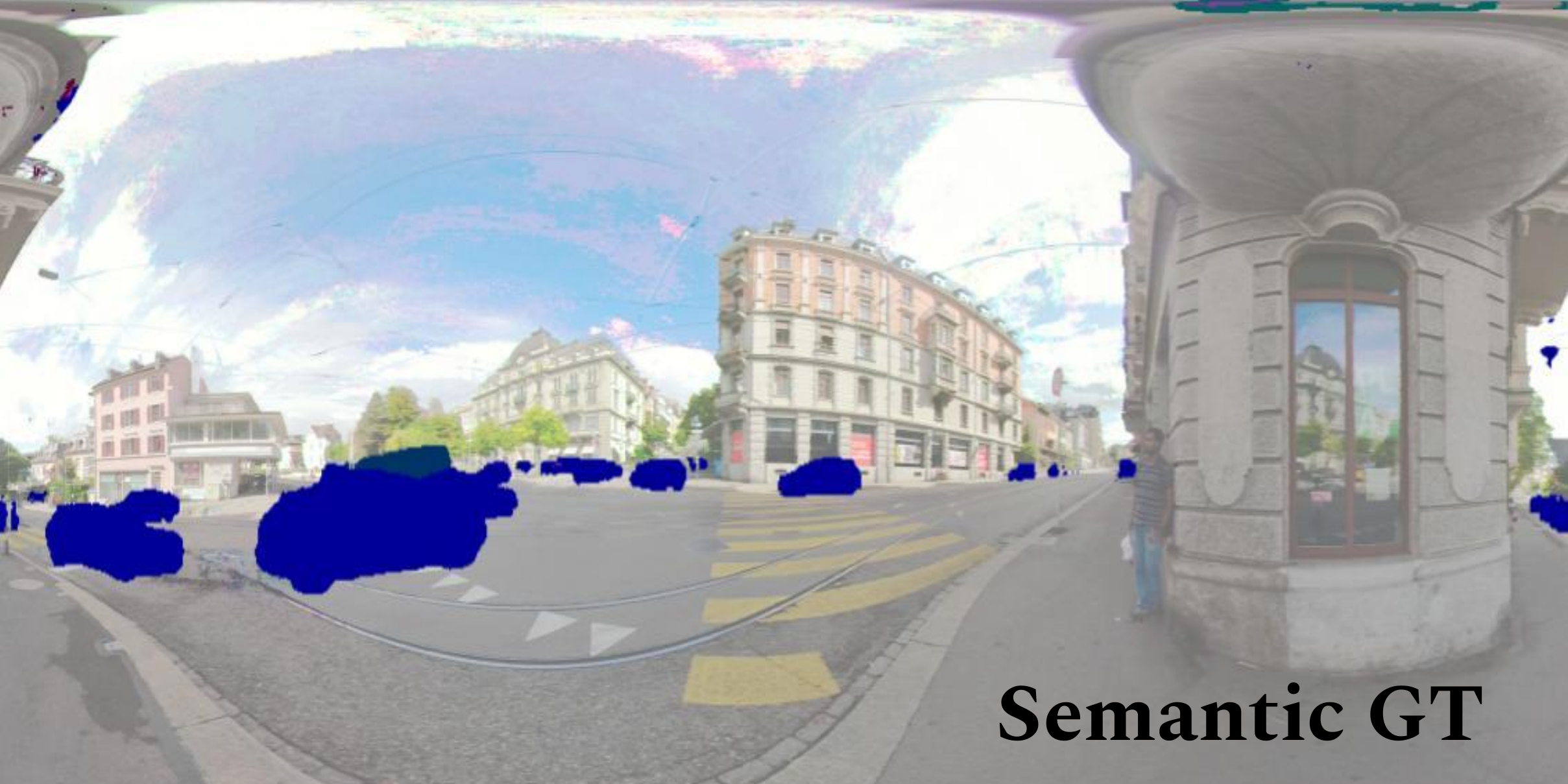} &
  \includegraphics[trim=0 0 0 0,clip,width=0.235\textwidth,height=0.12\textwidth]{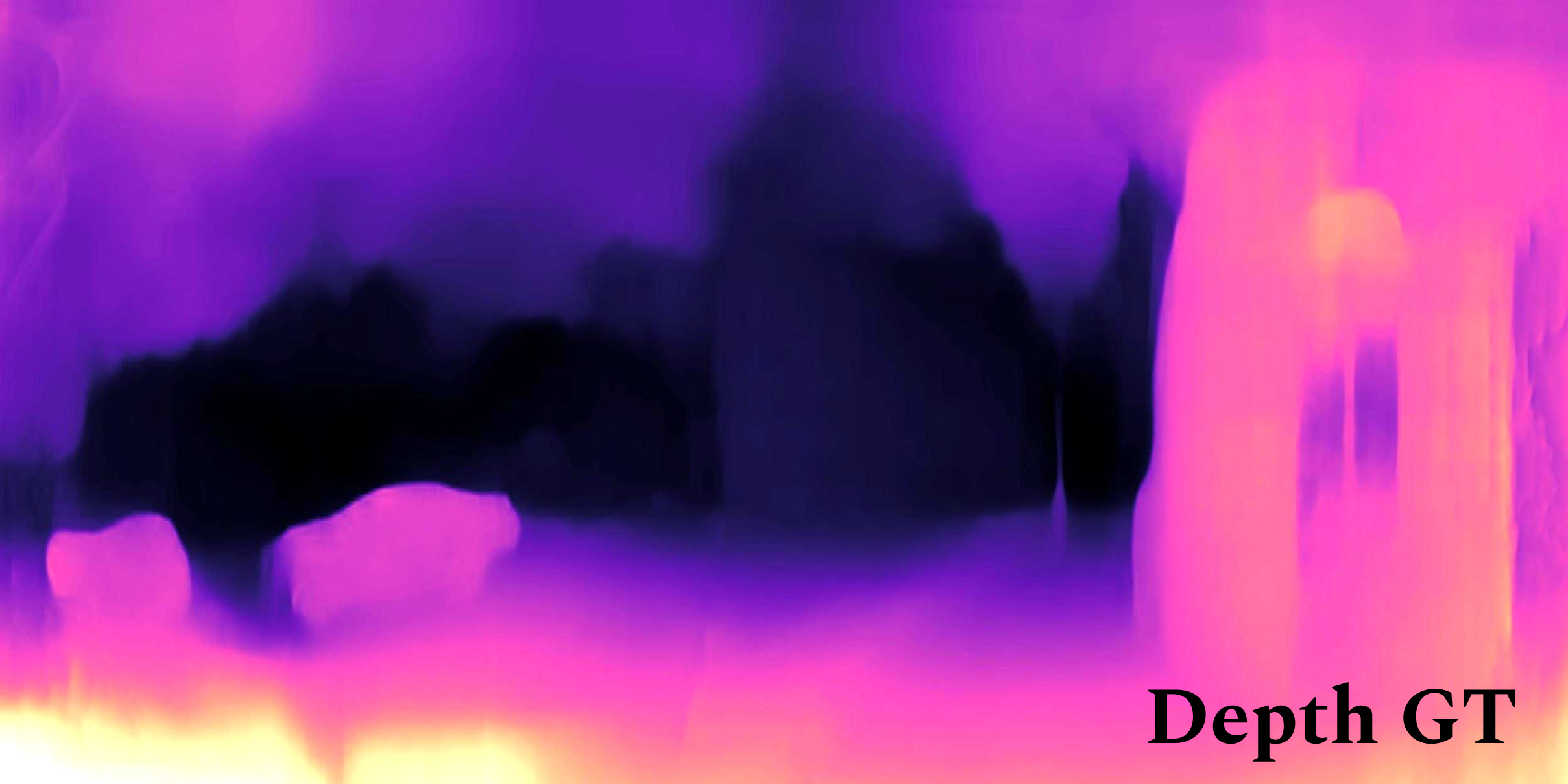} &
  \\
  \caption{Qualitative results of all three tasks. Better view in color.}
  \label{fig:multitask}\vspace{-5mm}
  \end{tabular}
\end{figure*}

\vspace{-2mm}
\subsection{Qualitative results}
We show qualitative results in Fig.~\ref{fig:result1} for the task of auditory semantic prediction. We also show the detected background image and the ground truth segmentation mask.
The last three rows are devoted to the results in 
rainy, foggy and night conditions respectively. We observe that our model remains robust to adverse visual condition. Of course, if the rain is too big, it will become an adverse auditory condition. In Fig.~\ref{fig:multitask}, we show two results by the multi-task setting. We show the predictions and the ground truths for all the three tasks.
It can be seen that the major sound-making objects can be properly detected. We see that the depth results reflect the general layout of the scene, though they are still coarser than the results of a vision system. This is valuable given the fact that binaural sounds are of very low-resolution -- two channels in total. More qualitative results are provided in the supplemental material.


\vspace{-2mm}
\subsection{Limitations and future work}
We obtain the ground truth of semantic segmentation and depth prediction by using the s-o-t-a pretrained vision models. These pretrained models remain as the upper bound in our evaluations. Moreover, the vision models are pretrained on perspective images and we apply them to panoramic images. This is due to the limited amount of datasets and annotations for panoramic images. We would like to note that most of the interesting objects appear in the equatorial region. For that region, the distortion of the panoramic images is less severe, and hence the results are less affected. In future, we plan to incorporate a 3D LiDAR to our sensor setup to get accurate ground-truth for depth prediction. 

We work with 3 object classes for semantic prediction task. This is because some classes are very rare in the middle-sized dataset, which already takes a great deal of effort to create. 
In comparison, the recent works~\cite{vehicle:tracking:sound:iccv19,irie2019seeing} only deals with one class in real world environment. To our best knowledge, we are the first to work with multiple classes in an unconstrained real environment.

\section{Conclusion} 
\label{sec:conclusion} 
The work develops an approach to predict the semantic labels of sound-making objects in a panoramic image frame, given binaural sounds of the scene alone. To enhance this task, two auxiliary tasks are proposed -- dense depth prediction of the scene and a novel task of spatial sound super-resolution. All the three tasks are also formulated as multi-task learning and is trained in an end-to-end fashion.
This work has also proposed a novel dataset Omni Auditory Perception dataset. Extensive experiments have shown that 1) the proposed method achieves promising results for all the three tasks; 2) the three tasks are mutually beneficial and 3) the number and orientations of microphones are both important.

%
%
\section*{Acknowledgement}
This work is funded by Toyota Motor Europe via the research project TRACE-Zurich. We would like to thank Danda Pani Paudel and Vaishakh Patil for helpful discussions.

\bibliographystyle{splncs04}
\bibliography{egbib}
\end{document}